\documentclass[runningheads]{llncs}

\usepackage[mobile]{eccv}

\usepackage{eccvabbrv}

\usepackage{graphicx}
\usepackage{nicematrix}
\usepackage[table]{xcolor}
\usepackage{xcolor}

\usepackage{soul}
\usepackage{makecell} %
\usepackage{todonotes}
\usepackage{booktabs}
\usepackage{multirow}
\usepackage{ulem}
\usepackage{soul}
\usepackage{enumitem}
\usepackage{amsmath}
\usepackage{wrapfig}
\usepackage{tikz}

\usepackage{tcolorbox}
\definecolor{lightyellow}{rgb}{1, 0.95, 0.85}

\newcommand{\methodcolor}[0]{green!20}
\newcommand{\findingcolor}[0]{green!15}
\newcommand{\model}{SteerViT}

\definecolor{clsbg}{RGB}{232,242,255}
\definecolor{segbg}{RGB}{232,248,238}
\definecolor{actbg}{RGB}{255,241,224}

\definecolor{clstext}{RGB}{30,90,170}
\definecolor{segtext}{RGB}{30,130,70}
\definecolor{acttext}{RGB}{190,105,20}

\definecolor{visionblue}{RGB}{214,233,253}
\definecolor{textred}{RGB}{248,214,201}
\definecolor{dullYellow}{RGB}{250, 240, 190}
\sethlcolor{\methodcolor}

\usepackage{siunitx}
\usepackage{threeparttable}
\usepackage{wasysym}
\usepackage{graphicx}
\usepackage{pifont} %
\newcommand{\cmark}{\ding{51}}
\newcommand{\xmark}{\ding{55}}

\usepackage[at]{easylist}

\usepackage{graphicx}   %
\usepackage{tabularx}
\usepackage{array}
\usepackage{booktabs}

\definecolor{trainbg}{RGB}{232,242,255}
\definecolor{inferbg}{RGB}{255,241,224}

\definecolor{traintext}{RGB}{30,90,170}
\definecolor{infertext}{RGB}{190,105,20}

\newcommand{\sceneheader}[1]{\textbf{#1}\par\vspace{1pt}}
\newcommand{\promptitem}[2]{\scriptsize\scriptsize\ttfamily ``#2''\par}

\makeatletter
\renewcommand\paragraph{\@startsection{paragraph}{4}{\z@}{0.2ex}{-1em}{\normalfont\normalsize\bfseries}}
\makeatother

\newcommand{\midsepremove}{\aboverulesep = 0.2mm \belowrulesep = 0.1mm}
\midsepremove
\newcommand{\midsepdefault}{\aboverulesep = 0.605mm \belowrulesep = 0.984mm}
\midsepdefault

\midsepremove
\newif\iffindingbox
\findingboxtrue

\usepackage[accsupp]{axessibility}  %

\usepackage{hyperref}

\usepackage{orcidlink}

\begin{document}

\title{
Steerable Visual Representations
}

\author{Jona Ruthardt\inst{1}\textsuperscript{*} \and
Manu Gaur\inst{2}\textsuperscript{*}\and \\
Deva Ramanan\inst{2} \and
Makarand Tapaswi\inst{3}\textsuperscript{$\dagger$} \and
Yuki M. Asano\inst{1}\textsuperscript{$\dagger$}}

\authorrunning{J.~Ruthardt and M.~Gaur et al.}

\institute{University of Technology Nuremberg \and
Carnegie Mellon University \and
International Institute of Information Technology, Hyderabad}

\maketitle

\begingroup
\renewcommand\thefootnote{}
\footnotetext{\textsuperscript{*} Equal contribution. \textsuperscript{$\dagger$} Equal advising.}
\endgroup

\vspace{-5mm}
\begin{abstract}
Pretrained Vision Transformers (ViTs) such as DINOv2 and MAE provide generic image features that can be applied to a variety of downstream tasks such as retrieval, classification, and segmentation.  
However, such representations tend to focus on the most salient visual cues in the image, with no way to direct them toward less prominent concepts of interest.  
In contrast, Multimodal LLMs can be guided with textual prompts,
but the resulting representations tend to be language-centric and lose their effectiveness for generic visual tasks.
To address this, we introduce
\textit{Steerable Visual Representations}, a new class of visual representations, whose global and local features can be steered with natural language.
While most vision-language models (e.g., CLIP) fuse text with visual features after encoding (\textit{late} fusion), we inject text directly into the layers of the visual encoder (\textit{early} fusion) via lightweight cross-attention. 
We introduce benchmarks for measuring \textit{representational steerability}, and demonstrate that our steerable visual features can focus on 
any desired object in an image while preserving the underlying representation quality.\textbf{}
Our method also matches or outperforms dedicated approaches
on anomaly detection and personalized object discrimination, exhibiting zero-shot generalization to out-of-distribution tasks.
\vspace{1mm}

\textbf{Project Website:} 
\href{https://jonaruthardt.github.io/project/SteerViT/}{jonaruthardt.github.io/project/SteerViT}
\vspace{-8mm}
\end{abstract}

\begin{figure}[h!]
\centering
\includegraphics[width=1.0\linewidth]{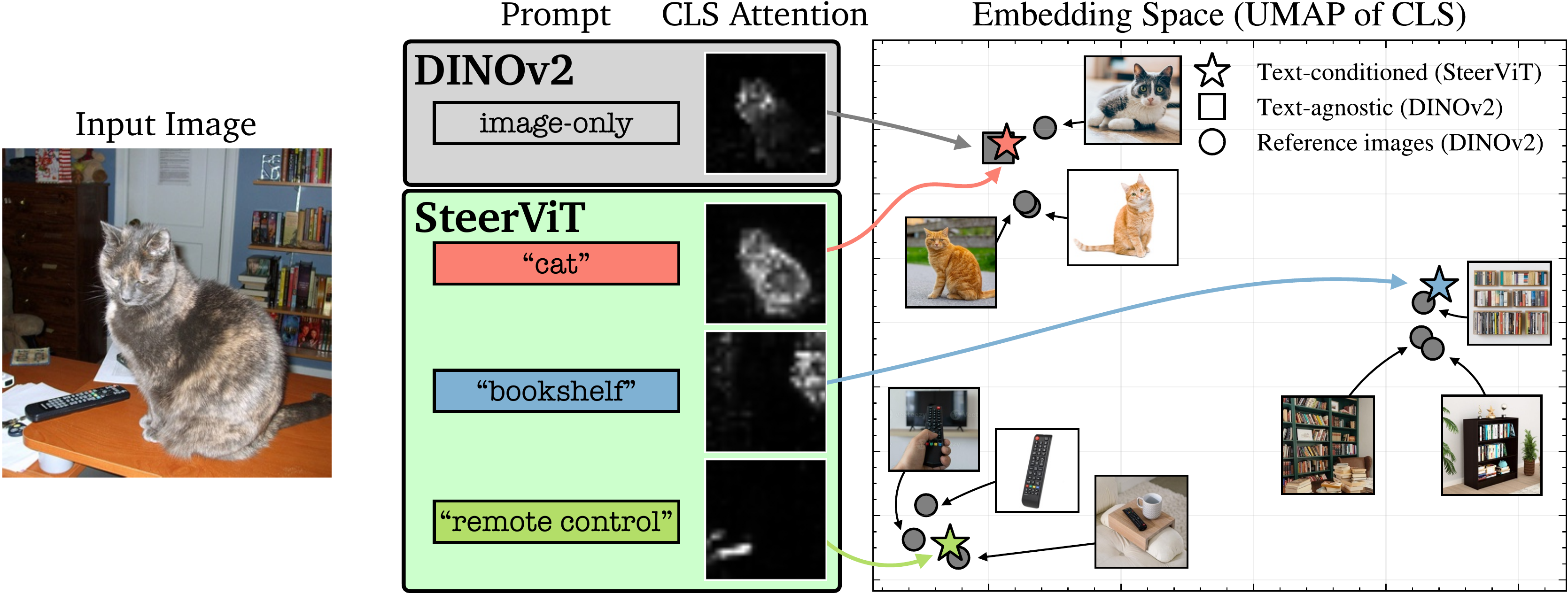}
\vspace{-6mm}
\caption{
\textbf{Steering visual representations with language.}
While 
{\setlength{\fboxsep}{2pt}\colorbox{gray!40}{DINOv2}} 
primarily encodes the salient object, producing a ``cat'' representation,
{\setlength{\fboxsep}{2pt}\colorbox{\methodcolor}\model{}}
can be steered with text to shift its attention (middle) and global feature semantics (right) towards the queried visual concept (e.g.,~``bookshelf'' or ``remote control'').
}
\vspace{-6mm}
\label{fig:heatmap_teaser}
\end{figure}

\begin{figure}[t]
\centering
\includegraphics[width=1.0\linewidth]{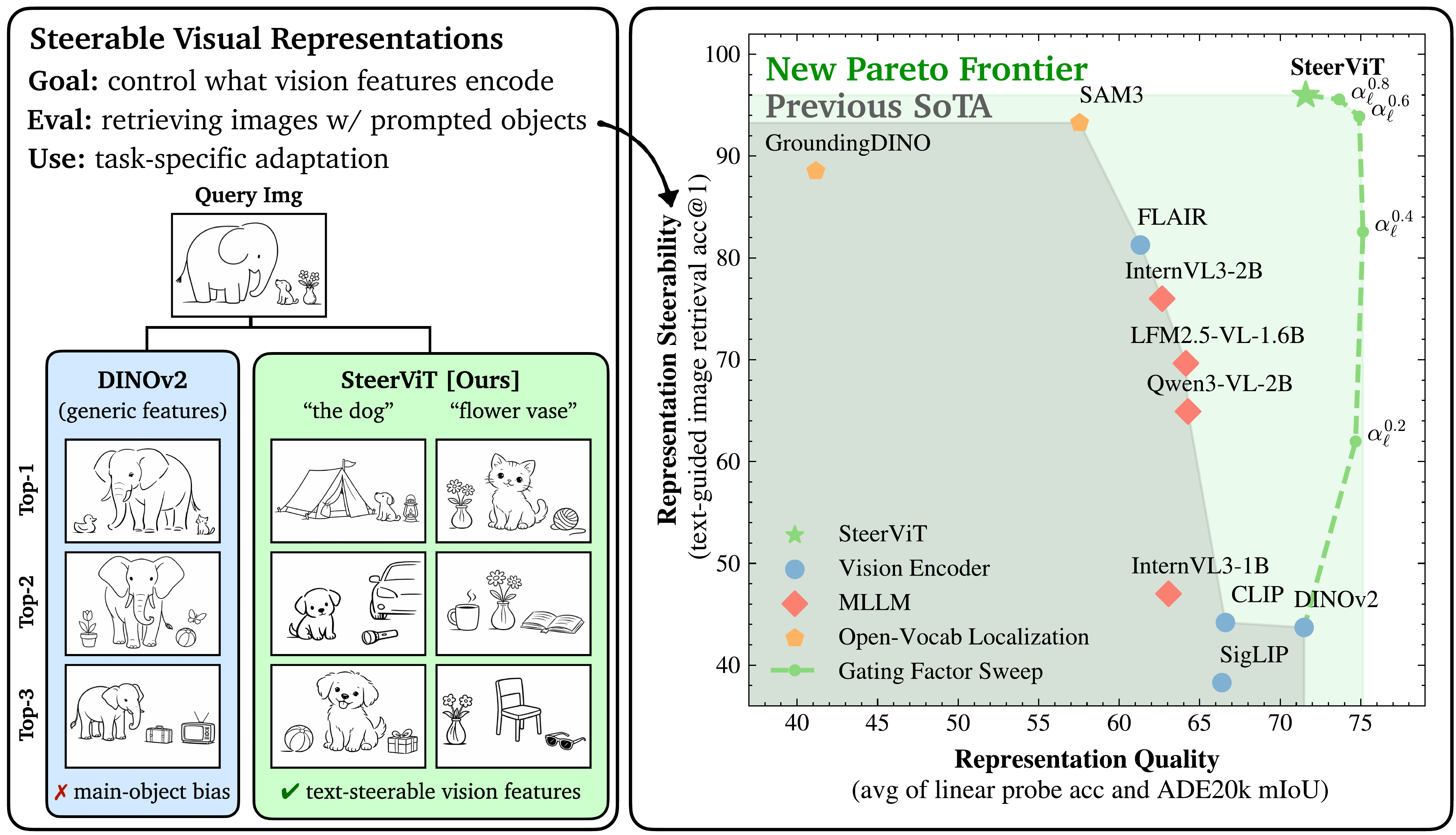}
\vspace{-4mm}
\caption{
\textbf{\model{} produces high-quality visual representations that can be steered by text.}
\textbf{Left:} Traditional (non-steerable) representations like {\setlength{\fboxsep}{2pt}\colorbox{visionblue}{DINOv2}} tend to focus on the dominant object in an image and retrieve images with the same object.
{\setlength{\fboxsep}{2pt}\colorbox{\methodcolor}\model{}} can adapt to a text prompt,
enabling retrieval of images even with small objects of interest.
\textbf{Right:} We compare \model{} to prior work in terms of its ability to adapt to text (measured by text-guided image retrieval (cf. ~\cref{subsec:cond_ret_exp})) and the quality of the visual representation (measured by the accuracy of 
linear probing for the CLS feature and semantic segmentation for patch features).
While models typically trade off steerability for representation quality, \model{} preserves both. By modulating a gating factor (\cref{eq:gated_ca}), \model{} achieves a new Pareto frontier.
\vspace{-4mm}
}
\label{fig:teaser}
\end{figure}

\section{Introduction}
\label{sec:intro}

Pretrained Vision Transformers (ViTs) such as DINOv2~\cite{dinov2}, MAE~\cite{he2022mae}, and SigLIP~\cite{zhai2023_siglip} provide a generic representation of an image that can be applied to a variety of downstream tasks such as retrieval, classification, and segmentation.
However, such representations tend to focus on the most prominent object in the image, likely due to the well-known photographer bias and object-centric vision datasets~\cite{torralba2011unbiased}.
Consider the indoor scene from \cref{fig:heatmap_teaser}: DINOv2 encodes the image focusing on the dominant salient object (``cat''), neglecting smaller or less prominent objects
like the ``remote control'' or ``bookshelf''.

Given the lack of other input, focusing on the dominant object is reasonable. %
However, tasks such as fine-grained localization may require greater consideration of less prominent visual concepts.
We argue that generic visual representations that can be \textit{steered} using task-specific priors have significant utility.

In this work, we seek to steer %
pretrained vision transformers with natural language.
We posit two desiderata that steerable representations should satisfy:
\begin{itemize}[nosep]
\item \textbf{Steerability:} 
The representation should adapt to the input text. In particular, it should be steerable 
in \textit{what} it encodes
(so that it captures objects irrespective of saliency;~\cref{subsec:cond_ret_exp})
and \textit{how} it organizes the embedding space
(so that clusters can be defined by specific attributes such as supercategories or object parts;~\cref{sec:tsne_viz}).

\item \textbf{Representation quality:} The steered representation should support diverse 
visual tasks, such as retrieval, classification, and segmentation (\cref{subsec:pareto}).
\end{itemize}

\begin{figure}[t]
\centering
\includegraphics[width=0.9\textwidth]{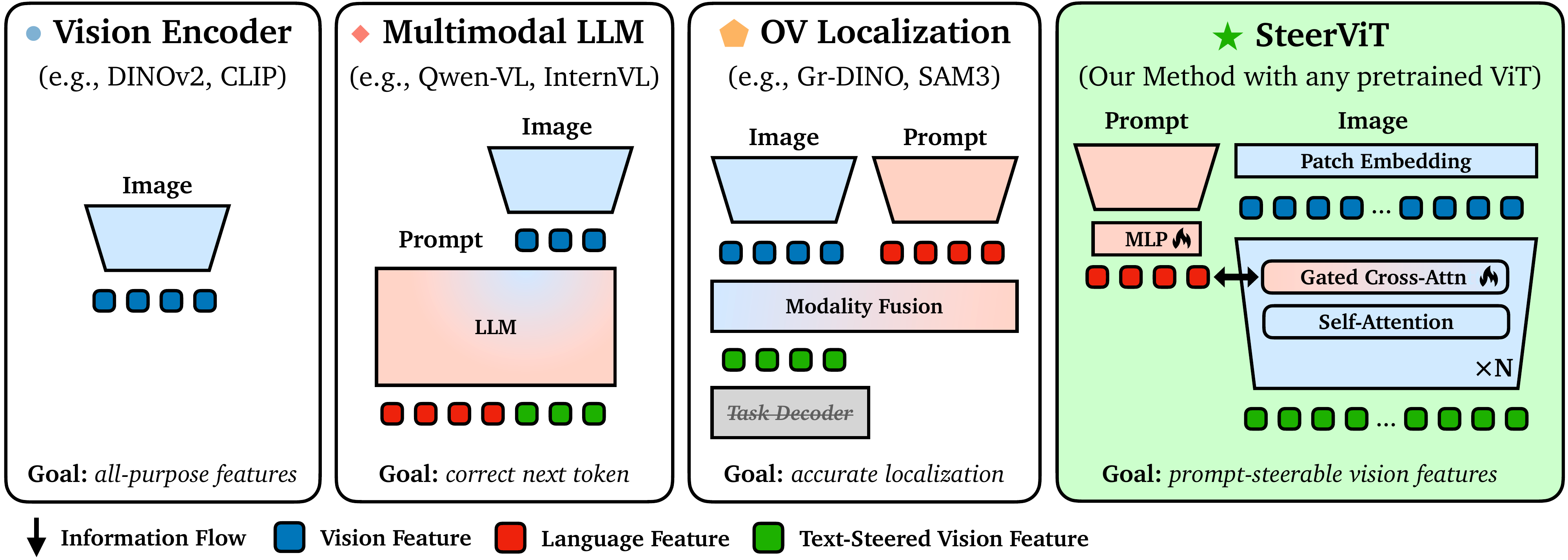}
\vspace{-3mm}
\caption{\textbf{Taxonomy of visual encoding.}
Standard vision encoders produce query-agnostic visual features.
MLLMs and OV Localization models \textit{late fuse} text after the visual encoder,
modeling vision-language interactions inside the LLM or task-aligned encoder.
\model{}, instead, directly steers the internal features of a frozen ViT using text prompts (\textit{early fusion}) via lightweight cross-attention layers. %
}
\vspace{-5mm}
\label{fig:arch_comparison}
\end{figure}

\noindent When considering prior art, we find that no existing approach simultaneously satisfies these desiderata (see \cref{tab:method_comparison}). Most vision-language architectures first encode images independently and only later fuse the modalities (see \cref{fig:arch_comparison}). 
As a result, text typically cannot influence the visual encoding process at inference time and can only operate on the vision encoder's fixed outputs. 
This stands in contrast to human vision, where prior textual priming can change how people parse images, often through top-down, task-guided attention~\cite{buswell1935people} (see \cref{apdx:motivation}).
While Multimodal LLMs (MLLMs) come closest to steerable vision-language representations, fusion in early layers of the language model typically yields language-dominant multimodal representations with diminished visual fidelity and limited controllability, as illustrated in \cref{fig:teaser}.

\begin{table}[t]
\centering
\footnotesize
\tabcolsep=1.4mm
\caption{
\textbf{Only \model{} satisfies both desiderata.}
\cmark=\! satisfied, \xmark~\!=\! not, \LEFTcircle~\!=\! partial.
\textit{Trainable MM Params} reflects 
multimodal
training; unimodal ViTs train solely on images.
\textit{V-L Fusion} notes where modalities interact relative to vision encoder layers. %
}
\vspace{-3mm}
\label{tab:method_comparison}
\resizebox{0.75\textwidth}{!}{
\begin{tabular}{l c c c c}
\toprule
Method & Text      & Feature & V-L    & Trainable \\
Family & Steerable & Quality & Fusion & MM Params \\
\midrule
Unimodal ViT (DINOv2) & \xmark & \cmark & \xmark & 0 \\
Cross-modal (CLIP) & \xmark & \cmark & Late & ${\sim}$200M \\
OV Localize (SAM3) & \cmark & \xmark & Late & ${\sim}$200M--1B \\
MLLM (Qwen3-VL) & \LEFTcircle & \LEFTcircle & Late, in LLM & ${\geq}$1B \\
\midrule
\model{} (Ours) & \cmark & \cmark & Early, in ViT & \textbf{21M} \\
\bottomrule
\end{tabular}
}
\vspace{-2mm}
\end{table}

With Steerable Visual Representations (\model{}), we \textit{invert} the MLLM paradigm and condition a visual encoder on language input, producing a \textit{vision-centric multimodal representation}.
Specifically, we interleave lightweight trainable cross-attention layers~\cite{alayrac2022flamingo} within frozen ViT blocks that attend to text prompts. 
This allows the visual encoding process to be heavily influenced or ``steered'' by text.
We adopt referential image segmentation as the training objective to encourage vision-language alignment.

Our method achieves a Pareto improvement over prior approaches (\cref{fig:teaser}),
producing visual features that are steerable with text while preserving their
underlying representation quality.
This is accomplished by freezing both the visual and text encoders and adding
only 21M trainable parameters via cross-attention, two orders of magnitude
fewer than MLLMs (see \cref{tab:method_comparison}).

Analogous to how prompting adapts (M)LLMs to novel tasks without retraining, language can adapt our \textit{steerable visual representations} to novel domains without fine-tuning.
Across diverse tasks spanning text-guided image retrieval (\cref{subsec:cond_ret_exp}), personalized object discrimination (\cref{subsec:text_density_exp}), and industrial anomaly detection (\cref{subsec:anomaly_detection}),
\model{}, without task-specific training, matches or outperforms strong baselines, including DINOv2, SAM3, billion-scale MLLMs and even some dedicated methods.
These results suggest that conditioning vision on language -- rather than language on vision -- could be a new paradigm for efficient multimodal vision understanding.

In summary, we make the following contributions:
\begin{enumerate}[nosep]
\item We introduce Steerable Visual Representations (SteerViT), a framework
that equips \textit{any} pretrained visual encoder with text-steerable representations via a simple grounding pretext task, adding only 21M parameters into the ViT.
\item We show that \model{} achieves a Pareto improvement over prior approaches, steering visual features with text while preserving feature quality.
\item We demonstrate that text prompts enable zero-shot generalization, allowing \model{} to transfer to new domains (e.g., personalized object discrimination or industrial anomaly detection) without task-specific training.
\end{enumerate}

\section{Related Work}
\label{sec:related_work}

\paragraph{Visual representation families.}
We summarize popular approaches against our desiderata in \cref{tab:method_comparison} and compare their architectures in \cref{fig:arch_comparison}.
Unimodal self-supervised encoders (DINOv2~\cite{dinov2}, MAE~\cite{he2022mae}) learn rich visual features but are query-agnostic.
Cross-modal encoders (CLIP~\cite{radford2021_clip}, SigLIP~\cite{zhai2023_siglip}) use text to provide training supervision; the visual encoder still cannot be steered with text.
MLLMs come closest with moderate steerability and visual quality, but their features reside in language space and require billions of parameters.
\model{} inverts this paradigm: we condition vision on language, add only ${\sim}$21M trainable parameters to a frozen ViT, while yielding stronger visual features (\cref{fig:teaser}).

\paragraph{Text-conditioned visual features.}

To our knowledge, \textit{no prior work} steers a visual encoder effectively with text while preserving its representation quality.
The closest attempt, FLAIR~\cite{xiao2025flair}, applies text-conditioned attention pooling over a frozen SigLIP encoder (late fusion), resulting in suboptimal steerability and underperforming unimodal encoders on standard vision benchmarks (\cref{fig:teaser}).
Concurrent works condition visual features on text but target narrow pipelines.
TIE~\cite{thirukovalluru2025tie} injects query tokens into the image encoder to reduce visual tokens in MLLMs, optimizing for document understanding.
ELIP~\cite{zhan2025elip} prepends text in the ViT to improve text-to-image retrieval re-ranking.
TEVI~\cite{mahajan2026tevi} edits CLIP's final image features via caption-conditional SAE masking to suppress irrelevant information.
In contrast, \model{} is a general framework for producing steerable visual representations that transfer across a wide variety of tasks.

\section{\model{}: Steering Vision Transformers with Text}
\label{sec:method}

\begin{wrapfigure}{r}{0.4\textwidth}
\centering
\vspace{-10mm}
\includegraphics[width=\linewidth]{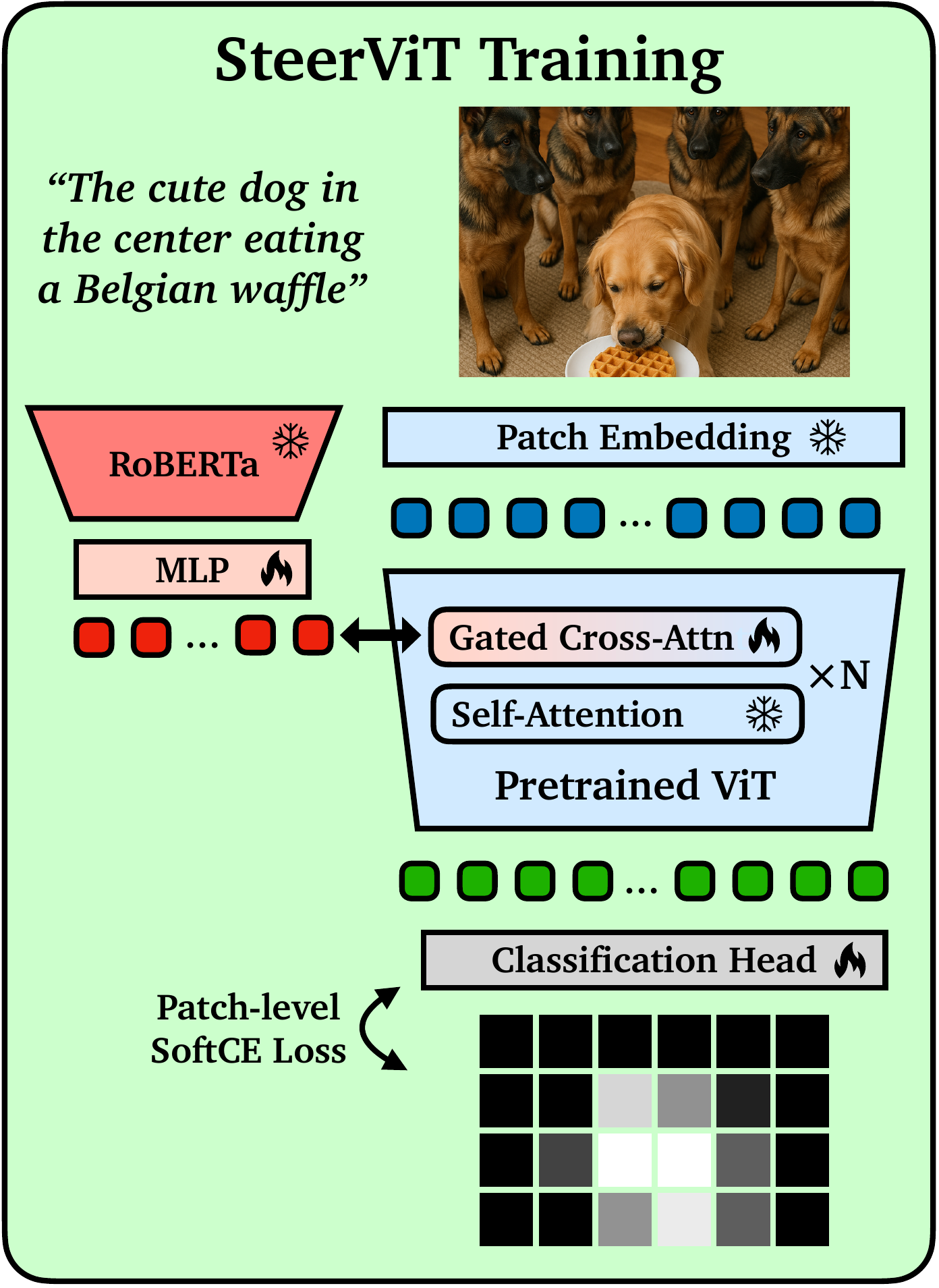}
\vspace{-6mm}
\caption{\textbf{Steering any ViT using text conditioning.}
Our method adds lightweight 
\colorbox{visionblue}{\textbf{vision}}-to-\colorbox{textred}{\textbf{language}} cross-attention layers within pretrained ViT blocks and applies a patch-level segmentation proxy objective to fuse prompt cues into patch tokens.
\label{fig:arch_training}}
\vspace{-12mm}
\end{wrapfigure}

\vspace{-2mm}
We describe the architectural modifications and post-training to obtain steerable representations from a pretrained ViT.

\vspace{-4mm}
\subsection{Architecture}
\vspace{-1mm}
\label{sec:arch}

As illustrated in \cref{fig:arch_training}, our \model{} consists of four components:

\colorbox{visionblue}{\textbf{A. Visual encoder.}}
For an image $X_v \in \mathbb{R}^{H \times W \times 3}$, a ViT produces a sequence of $N$ patch tokens $Z_v \in \mathbb{R}^{N \times d_v}$ and optionally a  \texttt{[CLS]} token ($d_v$ is the embedding dimension).
All original ViT parameters remain frozen throughout training and new capacity results exclusively through the interleaved cross-attention layers described below.
While most experiments adopt DINOv2 ViT-B/14~\cite{dinov2} as our backbone, we show that our approach also improves steerability of SigLIP and MAE.

\colorbox[RGB]{234,133,125}{\textbf{B. Text encoder.}}
We adopt a frozen, pretrained text encoder (RoBERTa-Large~\cite{liu2019roberta}) to produce token-level embeddings $Z_t \in \mathbb{R}^{L \times d_t}$ for a given input conditioning prompt $X_t$, where $L$ is a variable number of text tokens and $d_t$ is the text embedding dimension.

\colorbox{textred}{\textbf{C. Multimodal adapter.}}
Each text embedding $Z_t^i$ is $\ell_2$-normalized and fed through a trainable two-layer MLP that projects the sequence into a visual-aligned embedding space $H_t \in \mathbb{R}^{L \times d_v}$.

\tikz[baseline=(X.base)] 
\node[rectangle,
      inner sep=2pt,
      outer sep=0pt,
      shade,
      left color=textred,
      right color=visionblue] (X)
{\textbf{D. Gated cross-attention layers.}}; %
To fuse textual conditioning into the ViT's residual stream, we invert the gated cross-attention (CA) formulation from Flamingo~\cite{alayrac2022flamingo} by allowing hidden vision states to attend to language tokens (CA is language $\rightarrow$ vision in~\cite{alayrac2022flamingo}).
We insert CA layers into every other Transformer encoder block (e.g., 6 CA layers for 12 ViT-B blocks).
The visual patch tokens $Z_v^{(\ell)}$ at layer $\ell$ are queries and the adapted text tokens $H_t$ are keys and values:
{\small
\begin{equation}
\hat{Z}_v^{(\ell)} = \text{CA}(Z_v^{(\ell)}, H_t) = \text{softmax}\!\left(\frac{Q K^\top}{\sqrt{d_k}}\right) V,
\quad Q = Z_v^{(\ell)} W_Q, \; K = H_t W_K, \; V = H_t W_V .
\end{equation}
}
The output is integrated into the residual stream through a tanh gate with a layer-specific learnable scalar $\alpha_\ell$, initialized to zero:
\begin{equation}
Z_v^{(\ell+1)} = Z_v^{(\ell)} + \tanh(\alpha_\ell) \cdot \hat{Z}_v^{(\ell)}.
\label{eq:gated_ca}
\end{equation}
Since $\tanh(0) = 0$, the model is identical to the frozen ViT at initialization.
Despite the zero-initialization, the gate still receives a learning signal since
$\frac{\partial Z_v^{(\ell+1)}}{\partial \alpha_\ell}
=
\mathrm{sech}^2(\alpha_\ell) \cdot \hat{Z}_v^{(\ell)}$,
and $\mathrm{sech}^2(0)=1$. %
Thus, $\alpha_\ell$ can move away from zero during optimization, gradually activating the conditioning pathway and allowing the model to incorporate language-based clues.

\subsection{Training Objective}
\label{sec:objective}
In order to encourage the vision encoder to leverage and incorporate language clues, we design a pretext task requiring consideration of the prompt to be solved. We select referential segmentation for this purpose, where, given an image $X_v$ and a prompt $X_t$ referring to a target object or entity, the model predicts which patches correspond to the referred region.

The ground-truth $y_i$ is the fraction of foreground pixels of a pixel-wise binary segmentation mask that is patchified to match the ViT's $n \times n$ grid.
A linear classifier head maps each patch representation $Z_v^i \in \mathbb{R}^{d}$ to a mask probability $p_i$ through softmax.
We adopt the soft cross-entropy loss to train our model:
\begin{equation}
\textstyle
\mathcal{L} = - \sum_{i=1}^{n \times n} y_i \log p_i \, ,
\label{eq:loss}
\end{equation}
as it encourages the CA layers to route
textual information into the corresponding visual patch tokens, producing steered representations.
Performing segmentation on a patch- rather than pixel-level reduces the training complexity and forgoes the need for pixel-level decoders.

\subsection{Training Data}
\label{sec:data}
We train on a mixture of referential segmentation and grounding datasets spanning diverse visual domains and textual expression styles, comprising 162k unique images and 2.28M image-text pairs. Specifically, we use RefCOCO/+/g~\cite{kazemzadeh2014referitgame, yu2016refcocog}, Visual Genome~\cite{krishna2017visual}, LVIS~\cite{gupta2019lvis}, and Mapillary Vistas~\cite{neuhold2017mapillary}.
Details in \cref{apdx:data}.

\section{Experiments}
\label{sec:experiments}

In this section, we empirically validate properties of our learned steerable representations and show applications to diverse downstream tasks.%

\paragraph{Baselines.}
\label{subsec:baselines}

We compare \model{} (w/ DINOv2 ViT-B/14) against multiple model families that differ in how, and whether, they incorporate text into visual features (see \cref{app:baseline_extraction} for feature-extraction details):
(1)~\textbf{Unimodal vision encoders}: DINOv2~\cite{dinov2} and MAE~\cite{he2022mae} produce query-agnostic features with no text conditioning.
(2)~\textbf{Cross-modal encoders}: CLIP~\cite{radford2021_clip} and SigLIP~\cite{zhai2023_siglip}; we fuse visual and text embeddings via post-hoc element-wise addition (late fusion).
(3)~\textbf{MLLMs}: InternVL3~\cite{zhu2025internvl3}, Qwen3-VL~\cite{Qwen3-VL} and LFM-2.5-VL~\cite{amini2025lfm2technicalreport}; we extract prompt-specified vision features by following the last-token summary pooling of E5-V~\cite{Jiang2024E5VUE}.
(4)~\textbf{Open-vocabulary (OV) localization}: SAM3~\cite{carion2025_sam3} and GroundingDINO~\cite{liu2023groundingdino}; we use and evaluate the intermediate multimodal state.
Where supported, models process images at $336\times336$ resolution. %

\subsection{COnditional REtrieval: Steering Global Semantics with Text}
\label{subsec:cond_ret_exp}

\begin{figure}[t]
\centering
\begin{subfigure}[b]{0.38\textwidth}
\centering
\includegraphics[width=\textwidth]{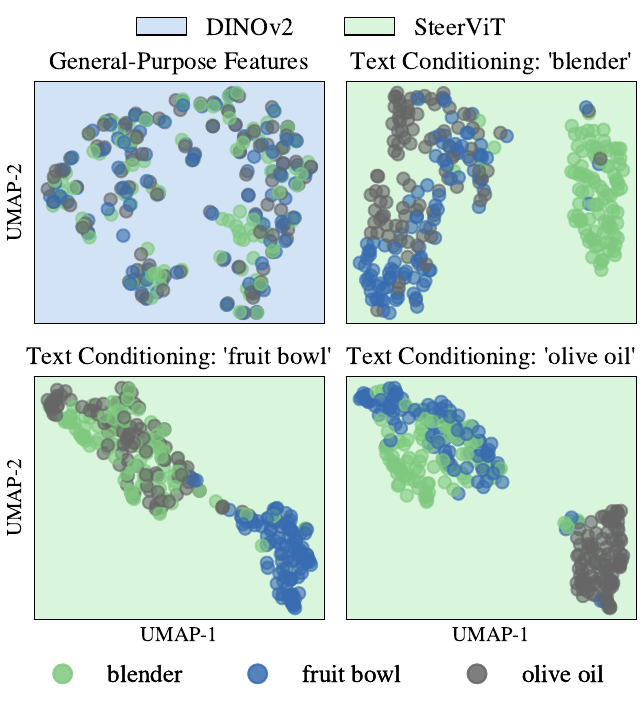}
\caption{
Feature distribution across three object types  in the ``kitchen'' scene.
}
\label{fig:umap-neighbourhood}
\end{subfigure}
\hfill
\begin{subfigure}[b]{0.61\textwidth}
\centering
\includegraphics[width=\textwidth]{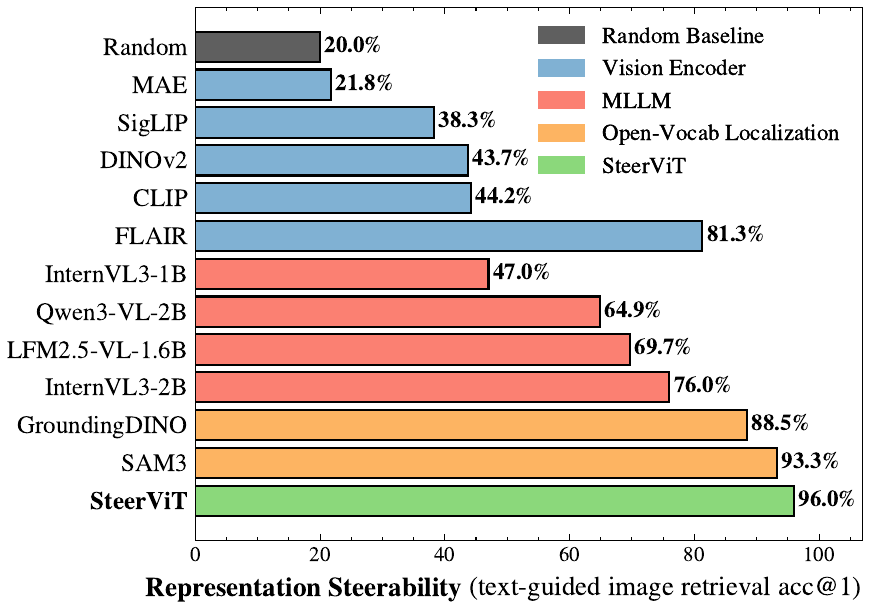}
\caption{
Non-salient object retrieval performance on CORE across model families.
}
\label{fig:cond_ret_bar}
\end{subfigure}
\caption{\textbf{COnditional REtrieval (CORE) benchmark.} \textbf{Left:} While {\setlength{\fboxsep}{2pt}\colorbox{visionblue}{DINOv2}} features form scene-level clusters, appropriate prompting of {\setlength{\fboxsep}{2pt}\colorbox{\methodcolor}\model{}} yields object-specific clusters. \textbf{Right:} Substantial differences in steerability between model families exist, with OV localization methods and \model{} offering the greatest adaptability. 
}
\vspace{-5mm}
\label{fig:cond_ret_baselines}
\end{figure}

We propose \textit{CORE (COnditional REtrieval)}, a text-conditioned image retrieval benchmark to measure how well a model steers its global features with text.

We select 100 images each for three indoor and three outdoor scenes from the SUN397 dataset~\cite{Herranz2016_SUNdataset} and inpaint 
five objects contextually fitting each scene into each image
using the FLUX.2 image editing model~\cite{flux-2-2025}
(e.g., a fruit bowl in a kitchen; a backpack in the park; see \ref{apdx:CORE} for details).
We frame the problem as one-vs-all retrieval: given a query image containing an inpainted object $\Omega$ in scene $\text{S}$ (e.g., a fruit bowl in a kitchen), the goal is to retrieve other images of $\text{S}$ that also contain $\Omega$.
This quantifies how well a model can steer its global features away from shared scene-level similarities (e.g., all images depict a kitchen) toward a specified non-salient object (e.g., the fruit bowl).
Both query and gallery images are encoded while conditioned on the same brief description of $\Omega$.
We measure the top-1 retrieval accuracy over the remaining 495 samples of $\text{S}$ with non-identical underlying images (pre-editing). %
Details about the experimental setup and results on a per-scene basis are reported in \ref{apdx:CORE}.

\paragraph{Query-agnostic encoders collapse to salient concepts.}

Query-agnostic encoders fail at conditional retrieval because their features collapse to the dominant scene concept (\cref{fig:cond_ret_baselines}).
MAE barely exceeds random chance ($20\%$) and DINOv2, despite its object-centric representations, achieves only 44\% acc@1.
Cross-modal visual encoders (CLIP, SigLIP) also perform poorly despite operating in a shared vision-language space.
In contrast, \model{} achieves 96\% retrieval accuracy, confirming that text conditioning shifts the global representation from the scene level (``kitchen'') to the queried concept (``fruit bowl''). 

\cref{fig:umap-neighbourhood} illustrates this on ``kitchen'' scenes for three objects via UMAP~\cite{UMAP}.
DINOv2 embeddings (top left) show no object separability,
whereas our text-conditioned embeddings form clusters for different text prompts.
This confirms that \model{} can reorganize its embedding space around the queried concept.

\paragraph{Late fusion does not enable steerability.}
Although CLIP and SigLIP operate in a shared vision-language space, their visual features are extracted independently of the query. Post-hoc element-wise addition of text yields a negligible 0.02\% boost over their vision-only representations, confirming that late fusion cannot steer frozen visual features.
In contrast, by conditioning intermediate representations on text (i.e., early fusion), \model{} improves retrieval accuracy from 43.7\% (vanilla DINOv2) to 96.0\%.
While FLAIR's trained attention pooling offers greater steerability (81.3\%) than typical cross-modal encoders, it still falls short of \model{} by 14.7 points.

\paragraph{MLLMs and OV models are steerable, but inefficient or specialized.}

MLLMs can moderately steer their visual features but at substantial computational cost.
\model{} outperforms both InternVL3-1B and InternVL3-2B by 49 and 20 percentage points, respectively, while only adding 21M parameters via cross-attention blocks compared to billion-parameter-scale LLMs (\cref{tab:method_comparison}).
Open-vocabulary localization models (GroundingDINO, SAM3) are highly steerable, with SAM3 nearly matching \model{}'s retrieval accuracy.
However, their intermediate representations are optimized for localization and lack the generality needed for downstream transfer, as discussed in \cref{subsec:pareto}.

\paragraph{Conditioning on a random class breaks retrieval.}

To verify that steerability is genuinely text-driven rather than an artifact of training, we condition each model on a random (incorrect) object class (\cref{tab:knn_exp_quant_full}).
Doing this has no effect on CLIP and SigLIP performance, confirming their features are unconditionally visual and not steerable.
However, performance degrades drastically for FLAIR and \model{} ($-29.4$ and $-48.3$ percentage points), indicating strong text-dependence and corroborating that steerability is primarily prompt-driven.
OV models see similar large drops, consistent with their deep text integration.
MLLMs show only mild sensitivity, with InternVL3-1B declining by $-7.6\%$.
\newline\newline
In summary, \model{} and OV localization models can reliably be steered through text whereas standard ViTs collapse to salient concepts, with post-hoc late fusion providing no benefits. Although MLLMs offer moderate steerability, they bear significant computational cost and diminished visual feature quality. 
Beyond CORE, \model{} also transfers to conditional retrieval on unaltered natural images in the GeneCIS benchmark~\cite{vaze2023gen}, reaching 25.4\% R@1 versus 9.6\% for DINOv2 and 18.7\% for the task-specialized baseline (details in \cref{apdx:genecis}).

\subsection{MOSAIC Localization: Text enables Targeted Attention}
\label{subsec:mosaic_exp}

We examine how \model{} routes and aggregates global representations via attention to query-relevant tokens.
For this, we construct a benchmark by stitching together four images from PASCAL-VOC \cite{pascal_voc} into a single $2\times2$ mosaic, resulting in a total of 363 composite images with reduced saliency of each primary subject.
The \texttt{[CLS]}-to-patch attention scores in the final self-attention block highlight how global attention is redirected to regions specified by text prompt. 

\begin{figure}[t]
\centering
\includegraphics[width=0.8\textwidth]{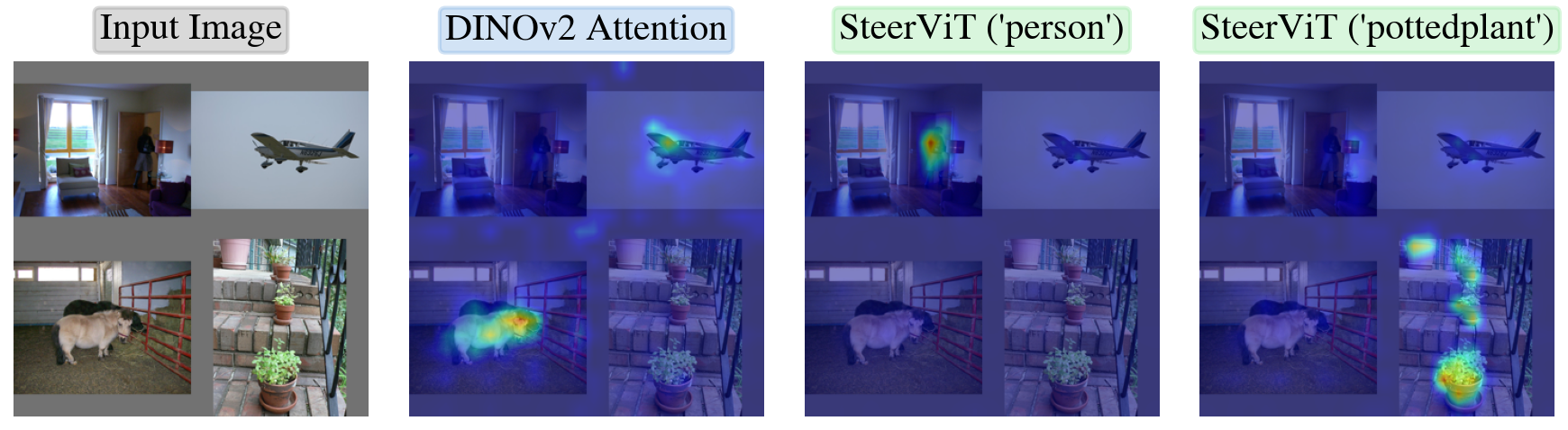}
\vspace{-3mm}
\caption{\textbf{Text enables targeted attention.} Attention maps on a four-image mosaic demonstrate that text conditioning with {\setlength{\fboxsep}{2pt}\colorbox{\methodcolor}\model{}} redirects self-attention to the queried concept whereas {\setlength{\fboxsep}{2pt}\colorbox{visionblue}{DINOv2}} attends to the most prominent objects. Note that the \texttt{[CLS]} token of \model{} was not directly optimized for targeted attention and remains frozen in its original state. 
}
\vspace{-4mm}
\label{fig:steer_attn}
\end{figure}

\paragraph{Qualitative analysis.}
As revealed by \cref{fig:steer_attn}, DINOv2's attention map focuses primarily on the prominent ``pony'' and ``airplane'' entities, confirming the saliency bias hypothesized in \cref{sec:intro}.
In contrast, \model{} conditioned on ``person'' routes attention to the barely visible person in the top-left.
When prompted with ``potted plant'', it distributes attention among the class instances in the bottom-right image. 
Additional qualitative examples are in \cref{apdx_fig:steer_attn}.

\paragraph{Quantitative evaluation.}
We quantify this effect by measuring the area under the precision-recall curve (PR-AUC) of the attention heatmaps and the ground-truth segmentation masks for each object type appearing in the mosaic instance.
DINOv2 cannot be actively steered and primarily focuses on prominent objects, resulting in a low PR-AUC of 14.3\%.
\model{} can be steered with text to focus on objects of interest, achieving a substantially higher score of 50.2\%.                                                                                                                      

\iffindingbox
\begin{tcolorbox}[colback=\findingcolor,
colframe=black, arc=4pt, boxsep=1pt]
\paragraph{\textbf{\textit{Finding 1}.}}
\model{} steers its visual features with text towards queried concepts, whereas standard encoders collapse to prominent visual cues.
\end{tcolorbox}
\vspace{-1.5em}
\fi

\subsection{%
Preserving Visual Representation Quality while Steering}
\label{subsec:pareto}
While previous sections revealed steerability for \model{} and OV localization, this is only useful if it preserves the downstream transfer, an important property of good visual representations.
A naïve solution that overwrites visual features with text yields perfect steerability but destroys the underlying representation.

To assess this trade-off, we contrast CORE performance (cf. \cref{subsec:cond_ret_exp}) with vision-centric downstream tasks intended to measure representational quality. 
In addition to training linear probes on global features across three fine-grained classification datasets (ImageWoof \cite{Howard_Imagewoof_2019}, Waterbirds \cite{Sagawa2020_waterbirds}, StanfordCars \cite{stanford_cars}),
binary object-of-interest segmentation on ADE20k \cite{zhou2017scene} gauges the dense encoding performance.
Here, promptable models are conditioned on the superclass (e.g., ``dog'' for ImageWoof) and the ADE20k class name, respectively. For Fig.~\ref{fig:teaser} and~\ref{fig:gate_factor}, both scores are averaged. 
Additional details are in supplement~\ref{apdx:feature_quality}.

\paragraph{Steerability $\leftrightarrow$ Representation quality trade-off.}
\cref{fig:teaser} (right) maps models along these two axes, revealing three regimes:
(1)~\textbf{Open-vocabulary localization methods} (SAM3 and GroundingDINO) achieve high steerability but produce localization-specific features that score poorly on generic vision tasks.
(2)~\textbf{MLLMs} perform reasonably on classification but falter on dense prediction tasks and require billions of parameters for moderate steerability.
(3)~\textbf{Query-agnostic encoders} (DINOv2, SigLIP) produce rich transferable features but cannot be steered.
\model{} bridges this gap, achieving
high steerability 
while fully preserving the representation quality of the underlying ViT.

\begin{wrapfigure}{r}{0.375\textwidth}
\centering
\vspace{-8mm}
\includegraphics[width=\linewidth]{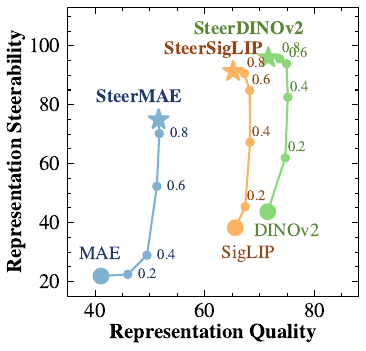}
\vspace{-6mm}
\caption{\textbf{Post-hoc modulation.} Scaling
CA gates $\alpha_\ell$ at inference for continuous interpolation between vanilla ViT and \model{} properties. A factor of $0.6$ provides the optimal steerability-quality trade-off for SigLIP- and DINOv2-based models.
\label{fig:gate_factor}}
\vspace{-6mm}
\end{wrapfigure}

\paragraph{Cross-attention gate as a continuous control knob.}
The tanh-gated cross-attention mechanism (\cref{eq:gated_ca}) provides an implicit control knob for
text conditioning strength.
At inference, we can scale the learned gating parameters $\alpha_\ell$ by a factor $\omega \in [0, 1]$ to smoothly interpolate between the unaltered ViT subspace and a fully text-conditioned state.
Plotting this trajectory (\cref{fig:teaser}, \cref{fig:gate_factor}) reveals a clear Pareto frontier with an optimal operating point at a scaling factor of $\omega {=} 0.6$ for DINOv2 and SigLIP, where both slightly exceed the original ViT's representation quality while unlocking high steerability.
Most strikingly, for MAE, 
representation quality monotonically improves as $\omega$ increases, rising from 40 at $\omega{=}0.0$ to 50 points at $\omega{=}0.6$.
Text conditioning enriches MAE's features with semantic structure, making them more transferable.

\iffindingbox
\begin{tcolorbox}[colback=\findingcolor,
colframe=black, arc=4pt, boxsep=1pt]
\paragraph{\textbf{\textit{Finding 2}.}}
\model{} produces the first family of visual representations that can be steered with text 
without sacrificing
the representation quality of the underlying vision encoder.
\end{tcolorbox}
\vspace{-1.5em}
\fi

\subsection{Text Specificity Guides Semantic Granularity}
\label{subsec:text_density_exp}
\begin{figure}[t]
\centering
\includegraphics[width=0.45\textwidth]{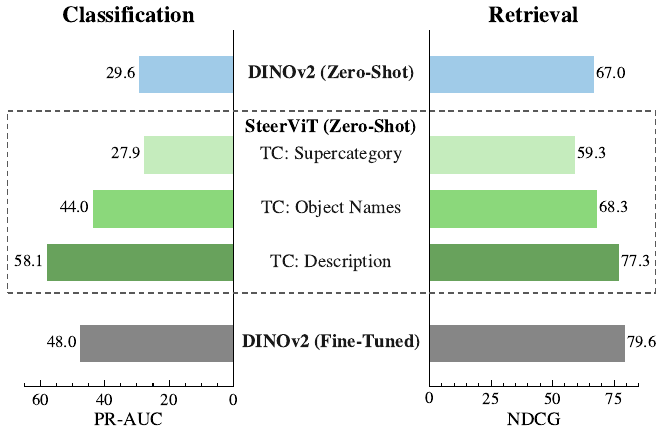}
\hfill
\includegraphics[width=0.48\textwidth]{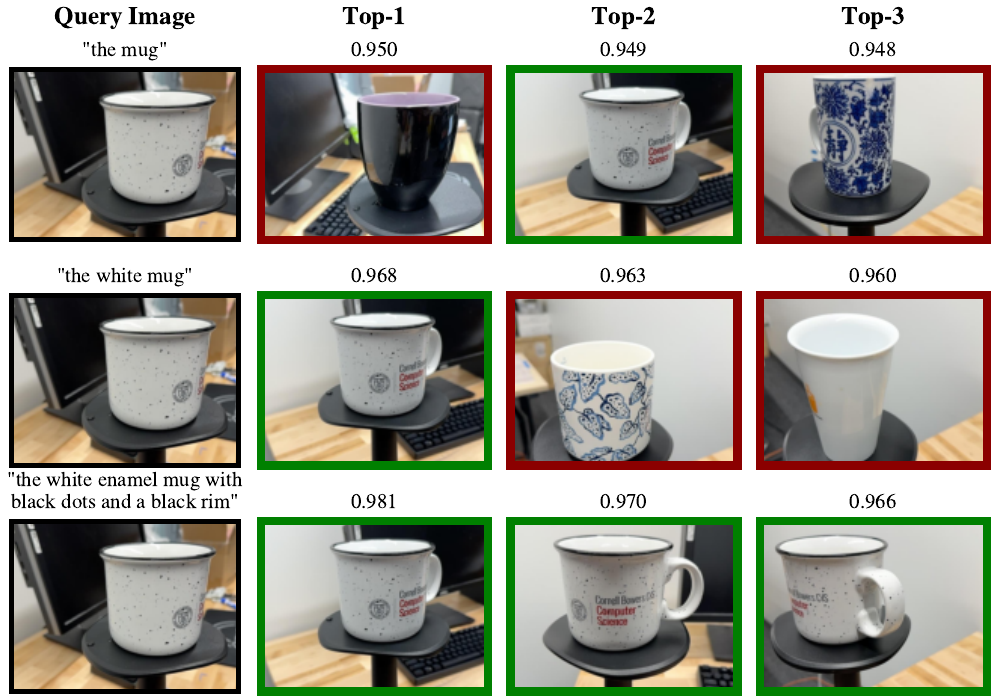}
\vspace{-2mm}
\caption{\textbf{Text controls feature granularity.} \textbf{Left:}
The quality of personalized representations produced by {\setlength{\fboxsep}{2pt}\colorbox{\methodcolor}\model{}} improves drastically with more detailed prompts for text conditioning (TC), even surpassing a supervised fine-tuned {\setlength{\fboxsep}{2pt}\colorbox{gray!75}{DINOv2}}.
\textbf{Right:} Retrieval improves when \model{} is given more detailed descriptions.
}
\vspace{-5mm}
\label{fig:pods}
\end{figure}

Next, we explore the role of text in forming high-fidelity visual representations.
For this, we choose Personal Object Discrimination Suite (PODS)~\cite{sundaram2024personalized_pods}, a benchmark evaluating the formation of instance-aware feature spaces via the ability of models to recognize particular objects (e.g., \textit{your} mug vs. all other mugs).
Given a small set of reference images, PODS computes cosine similarities on frozen global features and performs:
(i)~one-vs-all instance classification as well as
(ii)~similarity-based retrieval over a set of test images.
Following~\cite{sundaram2024personalized_pods}, we report PR-AUC for classification and NDCG for retrieval.

\paragraph{From coarse to fine-grained steering.}

We vary the amount of detail provided to \model{} via the text prompt from coarse supercategories (e.g., ``mug'', ``shoe''), object names (e.g., ``white ECCV mug''),
to comprehensive MLLM-generated descriptions of each instance's visual appearance. 
As shown in \cref{fig:pods}, the semantics of the steered representations are sensitive to the prompt specificity.
When conditioning is too coarse, the model overlooks fine-grained cues necessary for discriminating instances within an object category, yielding slightly worse performance than vanilla DINOv2 ($27.9\%$ vs.\ $29.6\%$ PR-AUC). Enriching the prompt with instance-level descriptions substantially boosts performance to $58.1\%$ PR-AUC, surpassing custom DINOv2 variants fine-tuned on synthetic task-specific data ($48.0\%$ PR-AUC) and nearly closing the gap 
between zero-shot and fine-tuned DINOv2 on 
retrieval ($77.3\%$ vs. $79.6\%$ NDCG).
This result is particularly noteworthy given that DINOv2 fine-tuning is object-specific, requiring a separate model per object class: 100 models in this setting compared to a single \model{} model.
These results demonstrate that \model{} does not simply \textit{add} information to the visual encoder but precisely controls the granularity of its visual features through the level of detail in the text prompt.

\iffindingbox
\begin{tcolorbox}[colback=\findingcolor,
colframe=black, arc=4pt, boxsep=1pt]
\paragraph{\textbf{\textit{Finding 3}.}}
The level of detail in text conditioning directly controls the granularity of the \textit{steerable visual representations}.
\end{tcolorbox}
\vspace{-1.5em}
\fi

\subsection{Visualizing and Analyzing the Steered Embedding Space}
\label{sec:tsne_viz}

\begin{figure}[t]
\centering
\includegraphics[width=\linewidth]{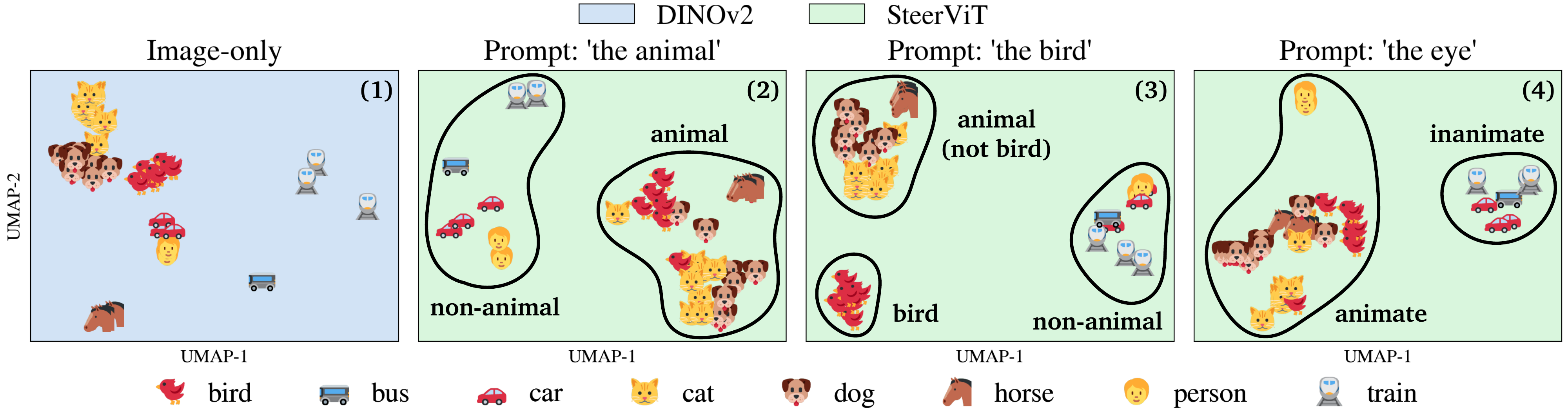}
\vspace{-6mm}
\caption{
\textbf{\model{} steers embedding topology via text.} We show sub-sampled UMAP projections of 500 images across eight PASCAL-VOC classes.
\textbf{(1)} {\setlength{\fboxsep}{2pt}\colorbox{visionblue}{DINOv2}} clusters by object class.
\textbf{(2)} Conditioning {\setlength{\fboxsep}{2pt}\colorbox{\methodcolor}\model{}} on ``animal'' merges animal classes into one cluster while separating non-animals.
\textbf{(3)} Conditioning on ``bird'' increases bird vs.\ non-bird separability while preserving the animal/non-animal macro-structure.
\textbf{(4)} Conditioning on ``eye'' groups all animate classes (including ``person'', clustered with inanimate classes previously) together, demonstrating compositional attribute steering.
}
\vspace{-4mm}
\label{fig:tsne_semantic}
\end{figure}

The results on PODS (\cref{subsec:text_density_exp}) indicate that query specificity dictates the level of semantic abstraction of \textit{steerable visual representations}.
Next, we explore how text conditioning shapes the embedding space topology of \model{} using two complementary modes: steering along the \textit{semantic hierarchy} and steering by \textit{compositional attributes}.

Here, we select 500 images containing exactly one out of eight curated classes from PASCAL-VOC~\cite{pascal_voc}. Encoded features are reduced to 2D using UMAP~\cite{UMAP}.

\paragraph{Steering Along the Semantic Hierarchy.}
As shown in \cref{fig:tsne_semantic}(1), DINOv2 embeddings cluster rigidly by object class.
Conditioned on ``animal'' (2), two macro-clusters emerge for \model{}, containing all animal and non-animal classes, respectively. Crucially, fine-grained structure is preserved (e.g., dogs and cats remain closer than birds), confirming that text controls clustering granularity without destroying object-level semantics.
When conditioned on ``bird'' (3) separability between bird and non-bird images increases while preserving the higher-level animal versus non-animal macro-separation.
This confirms that text can steer clustering at multiple levels of the semantic hierarchy, an emergent behavior not actively encouraged during training.

\paragraph{Steering by Compositional Attributes.}
Beyond semantic categories, text conditioning enables arbitrary grouping criteria, e.g., by shared object parts, defining a grouping principle orthogonal to semantic categories.
Conditioning on ``eye'' (\cref{fig:tsne_semantic}(4)) produces two macro clusters: animate classes that possess eyes versus inanimate objects that do not.
Notably, \textit{person} images, which were farther from animals under previous conditioning, now cluster together with them as the shared attribute ``has eyes'' overrides the semantic category boundary.
This demonstrates that we can steer representations by compositional properties, not just taxonomy.

\iffindingbox
\begin{tcolorbox}[colback=\findingcolor,
colframe=black, arc=4pt, boxsep=1pt]
\paragraph{\textbf{\textit{Finding 4}.}}
\model{} can reorganize its embedding space using text, controlling both the level of semantic abstraction and clustering criteria.
\end{tcolorbox}
\vspace{-1.5em}
\fi

\subsection{Text Facilitates Domain Transfer}
\label{subsec:anomaly_detection}

The previous sections showed that \model{} uses text to steer global features and route attention to queried objects, while preserving the representation quality of the underlying ViT. We now ask: \textit{Does this language-conditioned flexibility translate into generalization to novel downstream tasks and unseen domains?} 

\begin{figure}[t]
    \centering

    \begin{minipage}[t]{0.62\textwidth}
        \vspace{0pt} %
        \centering
        \includegraphics[width=\linewidth]{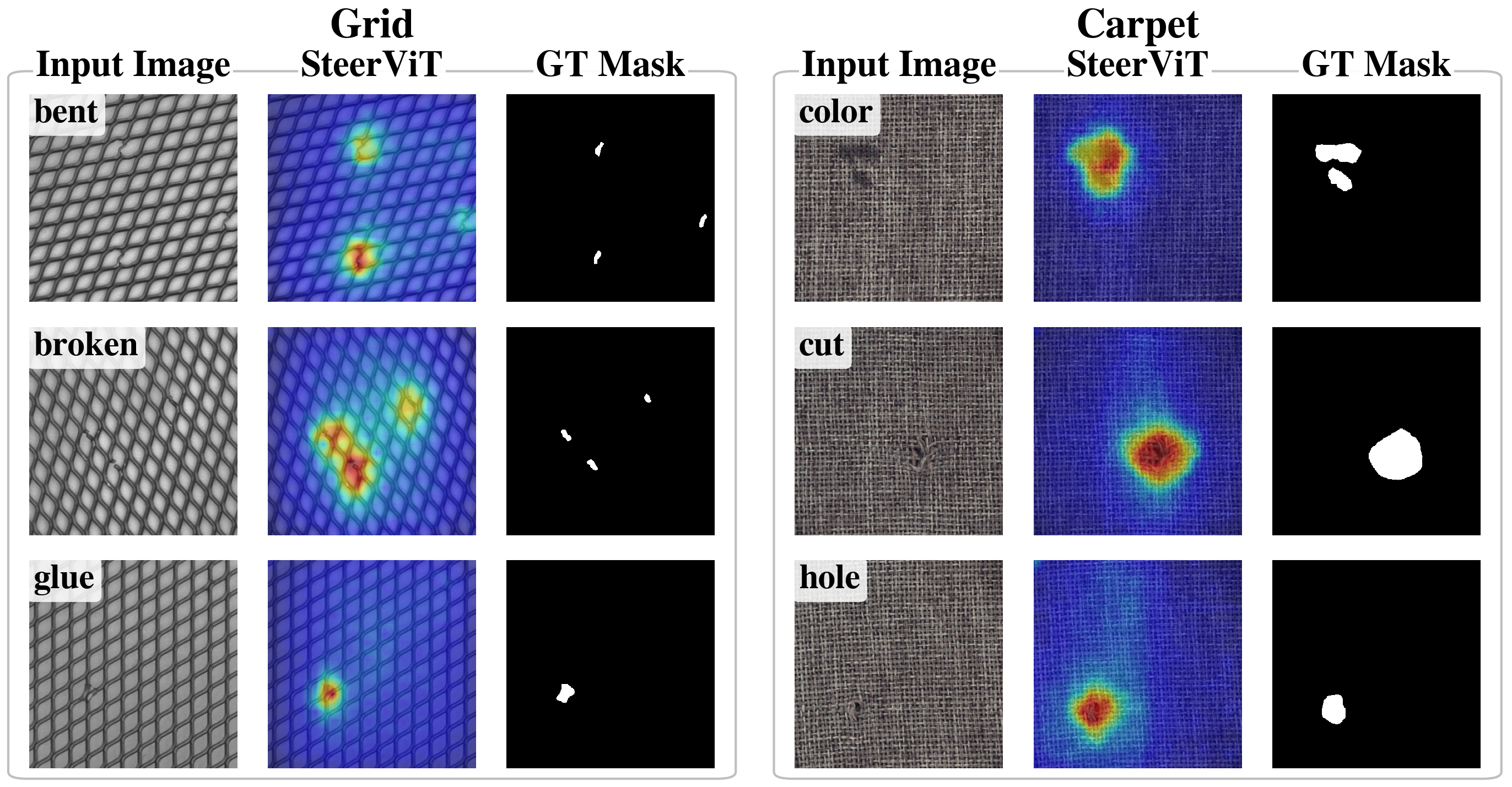}
        \vspace{-6mm}
        \caption{\textbf{Anomaly segmentation heatmap produced by \model{}.}
        Language-conditioning enables robust zero-shot generalization to this OOD task.}
        \label{fig:anomaly_seg_qualitative}
    \end{minipage}
    \hfill
    \begin{minipage}[t]{0.32\textwidth}
        \vspace{0pt} %
        \centering
        \captionsetup{type=table}
        \footnotesize
        \setlength{\tabcolsep}{4pt}
        \renewcommand{\arraystretch}{1.05}

        \begin{tabular}{@{}lc@{}}
        \toprule
        Method & PRO \\
        \midrule
        \textbf{MaskCLIP} \cite{zhou2022_maskclip} & 40.5 \\
        \textbf{CLIPseg} \cite{lueddecke22_clipseg} & 34.6 \\
        \textbf{SAM3} \cite{carion2025_sam3} & 54.5 \\
        \addlinespace
        \textcolor{gray}{\textbf{WinCLIP} \cite{WinCLIP_Jeong23CVPR}} & \textcolor{gray}{64.6} \\
        \textcolor{gray}{\textbf{DIVAD} \cite{hicsonmez2026training}} & \textcolor{gray}{73.3} \\
        \textcolor{gray}{\textbf{FADE} \cite{li2024_fade}} & \textcolor{gray}{84.5} \\
        \addlinespace
        \rowcolor{\methodcolor}\textbf{\model{}} & 82.1 \\
        \bottomrule
        \end{tabular}

        \caption{\textbf{ZS anomaly segmentation (MVTec).} 
        Dedicated methods in gray.}
        \label{tab:anomaly_detection_small}
    \end{minipage}
\end{figure}

Besides generalization to novel tasks like PODS (shown in Sec.~\ref{subsec:text_density_exp}), we showcase \model{}'s adaptability to extreme out-of-distribution settings by performing anomaly segmentation (AS) on the industrial MVTec~AD~\cite{bergmann_mvtech} dataset. %
Here, models are conditioned on prompts ``the anomaly in the <object>'' and anomaly maps are derived by upsampling the heatmaps produced by the learned linear segmentation head (\cref{fig:anomaly_seg_qualitative}).
We compare \model{} to other segmentation models and dedicated zero-shot AS methods using \textit{Per-Region-Overlap} (PRO).
More details are provided in \cref{apdx:anomaly_segmentation}.
\cref{tab:anomaly_detection_small} shows \model{} matching dedicated methods despite the extreme OOD setting.
These results reinforce the pattern of text-driven steering unlocking capabilities already present in the frozen ViT and transferring them to domains beyond the training distribution. 

\iffindingbox
\begin{tcolorbox}[colback=\findingcolor,
colframe=black, arc=4pt, boxsep=0.5pt]
\paragraph{\textbf{\textit{Finding 5}.}}
\model{} can use natural language to transfer rich unimodal vision encoders to OOD domains without task-specific training.
\end{tcolorbox}
\vspace{-1.5em}
\fi

\begin{wraptable}{r}{0.57\textwidth}
\centering
\footnotesize
\tabcolsep=0.5mm
\begin{threeparttable}
\vspace{-9mm}
\caption{%
We separately ablate \model{}'s three principal design choices (marked \xmark).
}
\vspace{-5mm}
\label{tab:arch_ablation}
\begin{tabular}{@{}ccc
                S[table-format=2.1]
                S[table-format=2.1]
                S[table-format=2.1]
                S[table-format=2.1]@{}}
\toprule
{Early} & {Tanh} & {MLP} & {FG-CLS} & {ADE} & {CORE} & {PODS} \\
{Fusion} & {Gate}  & {Proj.} & $\uparrow$ & $\uparrow$ & $\uparrow$ & $\uparrow$ \\
\midrule
\color{gray} {--}
  & \color{gray} {--}
  & \color{gray} {--}
  & \color{gray} 89.0
  & \color{gray} 53.7
  & \color{gray} 43.7
  & \color{gray} 29.6 \\
\rowcolor{\methodcolor}
\cmark & \cmark & \cmark
  & 87.7
  & 55.4
  & \textbf{96.0}
  & \textbf{58.1} \\
\addlinespace
\xmark & \cmark & \cmark
  & \textbf{91.8}
  & \textbf{55.5}
  & 93.3
  & 36.6 \\
\cmark & \xmark & \cmark
  & 83.5
  & 55.3
  & 94.6
  & 47.1 \\
\cmark & \cmark & \xmark
  & 86.7
  & 54.5
  & 95.2
  & 56.4 \\
\bottomrule
\end{tabular}
\xmark\ in Early Fusion means late fusion. %
\xmark\ in Tanh Gate means ungated cross-attention.
\xmark\ in MLP Proj.\ means single linear layer. %
Vanilla DINOv2 (top, gray) is vision-only.
\end{threeparttable}
\vspace{-6mm}
\end{wraptable}

\subsection{Analysis of \model{}}
\label{sec:ablations}

We ablate three key architectural and training decisions in \model{} to isolate the contribution of each. More ablations are in \cref{apdx:more_ablations}.
Unless otherwise noted, experiments use the DINOv2 ViT-B/14 backbone and the RoBERTa-Large language model and are trained at $336^2$ resolution with a batch size of 12 for 500k iterations (${\sim}$84 H100 GPU-hours) on the full data mixture (\cref{sec:data}), using AdamW with a cosine schedule that warms up to $3{\times}10^{-4}$ over 5k steps, decays to $3{\times}10^{-5}$ by 40k steps, and remains constant thereafter.
We test fine-grained classification probe accuracy (FG-CLS; \cref{subsec:pareto}), referential segmentation performance (ADE; \cref{subsec:pareto}), steerability via CORE (\cref{subsec:cond_ret_exp}), and PR-AUC on PODS with descriptive prompts (\cref{subsec:text_density_exp}). %

\paragraph{Architecture Choices.}
\label{subsec:arch_ablation}

\model{} has three principal design choices:
early fusion of text within the ViT,
tanh-gated CA layers, and
MLP text projector.
\cref{tab:arch_ablation} ablates each component.

\textit{Early vs.\ late fusion.}
We interleave gated cross-attention within ViT layers (early fusion).  
Late fusion (\cref{tab:arch_ablation}, row 3) adds text only after the final ViT layer.  
While late fusion (row 3) also achieves high steerability (93.3) and higher FG-CLS (91.8 vs.\ 87.7), it reduces PODS performance dramatically (36.6 vs.\ 58.1).
This illustrates the importance of early fusion for fine-grained tasks as the gap vanishes (26.5 vs. 27.9) when conditioning on coarse supercategory prompts.

\textit{Role of gating.}
Removing the zero-initialized tanh gate (row 4) reduces FG-CLS, CORE, and PODS by 4.2, 1.4, and 11.0 points below \model{} (row 2), showing that ungated cross-attention disrupts frozen features.
         
\textit{Language projector.}
A two-layer MLP projects RoBERTa features to the ViT space (\cref{sec:arch}).  
Replacing it with a linear projector (row 5) lowers FG-CLS by 1.0 and PODS by 1.7 points, indicating that the MLP better aligns modalities.

\paragraph{Generalization Across Pretrained Backbones.}
\label{subsec:cross_backbone}

\begin{wraptable}{r}{0.42\textwidth}
\centering
\footnotesize
\tabcolsep=1.2mm
\vspace{-12mm}
\caption{%
\textbf{Steering different ViT pretraining approaches.}
\model{} consistently outperforms late fusion, with the largest gains on weaker backbones.
}
\label{tab:cross_backbone}
\begin{tabular}{l cc >{\columncolor{\methodcolor}}c}
\toprule
\multirow{2}{*}{Backbone} & \multicolumn{3}{c}{CORE $\uparrow$} \\
& Base & Late & \model{} \\
\midrule
DINOv2 & 43.7 & 93.3 & \textbf{96.0} \\
SigLIP & 38.3 & 75.4 & \textbf{91.3} \\
MAE    & 21.8 & 41.0 & \textbf{74.9} \\
\bottomrule
\end{tabular}
\vspace{-8mm}
\end{wraptable}

To validate that our approach generalizes beyond DINOv2, we apply \model{} to two additional ViTs: SigLIP~\cite{zhai2023_siglip}
and MAE~\cite{he2022mae}.
\cref{tab:cross_backbone} compares vanilla, late fusion, and early fusion (\model{}) variants. %

Overall, DINOv2 produces the strongest steerable representations, consistent with its richer self-supervised features. SigLIP and MAE similarly benefit from text conditioning but start from weaker baselines.
Early fusion consistently outperforms late fusion across all three ViT families.
The advantage is more pronounced for MAE and SigLIP, where early fusion boosts steerability by 33.9 and 15.9 points, respectively. 
These larger early-fusion gains compared to DINOv2 (2.7) align with our layer-wise analysis (\cref{fig:feature_divergence}), where \model{} representations diverge earlier from their base ViTs for MAE and SigLIP, suggesting that language injection is beneficial when the underlying visual features are less semantically mature.

\section{Conclusion}

We introduce Steerable Visual Representations (\model{}), a new class of visual representations that equip \textit{any} pretrained visual encoder with 
the ability to steer its features with natural language. 
\model{} inverts the 
MLLM
paradigm by conditioning the visual encoder on language.             
By interleaving lightweight cross-attention layers into frozen ViT blocks, our method steers both global and local features with text while preserving the base ViT's representation quality, achieving a Pareto improvement over prior approaches with only ${\sim}$21M trainable parameters.
Across personalized object discrimination and industrial anomaly segmentation, 
our method
matches or surpasses 
dedicated methods without task-specific training, generalizing across multiple ViT backbones.
These results suggest that text can serve as a lightweight, post-hoc steering mechanism that extends the capabilities of rich 
vision encoders to new domains without fine-tuning.

\section*{Acknowledgements}
The authors gratefully acknowledge the HPC resources provided by the Erlangen National High Performance Computing Center (NHR@FAU) of the Friedrich-Alexander Universität Erlangen-Nürnberg (FAU) under the BayernKI project v115be. BayernKI funding is provided by Bavarian state authorities. The authors also gratefully acknowledge the Gauss Centre for Supercomputing e.V. (www.gauss-centre.eu) for funding this project by providing computing time on the Supercomputer JUPITER at Jülich Supercomputing Centre (JSC). MT thanks Adobe Research for a research gift.

\bibliographystyle{splncs04}
\bibliography{bib/longstrings, bib/main}

\newpage
\appendix
\appendix
\setcounter{secnumdepth}{3}   %
\setcounter{tocdepth}{3}      %

\makeatletter
\renewcommand\subsubsection{\@startsection{subsubsection}{3}{\z@}%
  {-1.25ex\@plus -0.5ex \@minus -0.2ex}%
  {0.8ex \@plus 0.2ex}%
  {\normalfont\normalsize\bfseries}}
\makeatother

\begin{center}
\textbf{\Large{Steerable Visual Representations}} \\
\textsc{Supplementary Material}
\end{center}

In the following, we discuss the inspiration of our work from human image perception (\cref{apdx:motivation}),
additional qualitative and quantitative results on CORE, GeneCIS, MOSAIC, and anomaly segmentation (\cref{apdx:more_results}),
additional ablations (\cref{apdx:more_ablations}), and
details of experimental setup such as training dataset and feature extraction process (\cref{apdx:exp_details}).

\section{Motivation: Parallels to Human Image Perception}
\label{apdx:motivation}

\begin{figure}[h!]
\centering
\includegraphics[width=0.4\textwidth]{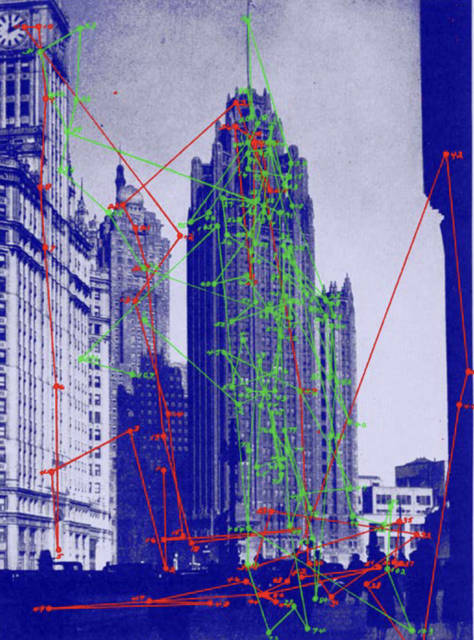}
\caption{\textbf{Human gaze patterns.} Eye movement patterns differ when observers are instructed to \textit{locate a person looking out of a window in the tower} (green) versus when no contextual instruction is provided (red).
Adapted from~\cite{buswell1935people}.
}
\label{fig:gaze_cogsci}
\vspace{-4mm}
\end{figure}

\cref{fig:gaze_cogsci} illustrates how human gaze patterns over the same image shift depending on a task-relevant textual prompt~\cite{buswell1935people}, providing inspiration from neuroscience for \model{}'s early fusion of language into the visual encoding process.

\section{Additional Qualitative and Quantitative Results}
\label{apdx:more_results}

This section provides further details about the employed evaluations, provides qualitative examples and reports additional results. 

\subsection{Assessing Representational Steerability}

\subsubsection{Controlled Prompt-Conditional Retrieval on CORE}
\label{apdx:CORE}

For our \textit{COnditional REtrieval} (CORE) benchmark introduced in \cref{subsec:cond_ret_exp}, we select 100 base images for each of six scenes (three indoor and three outdoor environments) from SUN397~\cite{Herranz2016_SUNdataset}. For each image, we create five separate instances by inpainting a scene-relevant background object (e.g., an ``olive oil bottle'' in ``kitchen'') using FLUX.2 [dev] \cite{flux-2-2025}, resulting in a total of 500 images per scene. 
The selected scenes, the corresponding five objects added to each scene and the text-conditioning prompts are reported in \cref{tab:core_object_prompts}.

To test the ability of visual encoders to steer their global representations in order to retrieve images relevant to a specified prompt, we frame the problem as one-vs-all retrieval. 
For each image $X_v^\Omega$ containing object $\Omega$, the objective is to retrieve other instances including $\Omega$. Here, all images (query and gallery instances) are encoded with the same object-specific prompt $X_t^\Omega$ (cf.~\cref{tab:core_object_prompts}). For a given query image, the gallery is comprised of 495 variations with non-identical base images of the same scene. The steerability score is calculated using top-1 retrieval accuracy and averaged across query images and object types.

\begin{table*}[t]
\caption{\textbf{Overview of scene and object types used in CORE benchmark.} Each of the three indoor and outdoor environments is paired with five corresponding objects, that are inpainted into images of that scene using an image editing model. The prompts provide contextual guidance, indicating the object the model should focus on.}
\label{tab:core_object_prompts}
\centering
\footnotesize
\setlength{\tabcolsep}{4pt}
\renewcommand{\arraystretch}{1.02}
\begin{tabularx}{\textwidth}{
    >{\raggedright\arraybackslash}X
    >{\raggedright\arraybackslash}X
    >{\raggedright\arraybackslash}X
}
\toprule
\multicolumn{3}{l}{\textbf{Indoor}} \\
\midrule
\sceneheader{Kitchen}
\promptitem{Blender}{the black blender with a transparent glass cup.}
\promptitem{Fruit Bowl}{the fruit bowl with fresh red apples.}
\promptitem{Olive Oil}{the green glass bottle of olive oil.}
\promptitem{Green Towel}{the dark green kitchen towel.}
\promptitem{Pizza}{the gray plate with a slice of pepperoni pizza.}
&
\sceneheader{Living Room}
\promptitem{TV Remote}{the small black TV remote.}
\promptitem{Roses}{the small bouquet of red roses in a white and gray vase.}
\promptitem{Books}{the stack of multiple books.}
\promptitem{Ball}{the black and white soccer ball.}
\promptitem{Wine Bottle}{the bottle of red wine.}
&
\sceneheader{Bathroom}
\promptitem{Rubber Duck}{the small yellow rubber duck.}
\promptitem{Mop}{the wet mop with a red handle.}
\promptitem{Toilet Paper}{the stack of white toilet paper rolls on the floor.}
\promptitem{Carpet}{the small pink carpet on the floor.}
\promptitem{Coat}{the hanging winter coat.}
\\
\midrule
\multicolumn{3}{l}{\textbf{Outdoor}} \\
\midrule
\sceneheader{Street}
\promptitem{Tree}{the apple tree with ripe apples.}
\promptitem{Trash Can}{the small metal trash can.}
\promptitem{Flag}{the American flag.}
\promptitem{Zebra}{the zebra walking on the street.}
\promptitem{Puddle}{the puddle with water on the street.}
&
\sceneheader{\shortstack[l]{Suburb}}
\promptitem{Garden Hose}{the coiled green garden hose.}
\promptitem{Mailbox}{the blue curbside mailbox.}
\promptitem{Bicycle}{the bicycle leaning against something.}
\promptitem{Basketball}{the orange basketball.}
\promptitem{Fire Hydrant}{the red fire hydrant.}
&
\sceneheader{Park}
\promptitem{Stroller}{the baby stroller.}
\promptitem{Traffic Cone}{the orange traffic cone.}
\promptitem{Birdhouse}{the small birdhouse on a post.}
\promptitem{Backpack}{the dark blue hiking backpack.}
\promptitem{Kite}{the colorful kite lying on the grass.}
\\
\bottomrule
\end{tabularx}
\end{table*}

We report detailed quantitative results on a per-scene basis in Table \ref{tab:knn_exp_quant_full}. 

\model{} (green) consistently outperforms query-agnostic ViTs (blue), MLLMs (red), and OV localization models (yellow) across all indoor and outdoor scenes, surpassing the underlying DINOv2 by an average of 52.5 points.
When conditioned on a randomly chosen incorrect object (\xmark), \model{}'s accuracy drops drastically, whereas cross-modal encoders (CLIP and SigLIP) see negligible changes.
This confirms that text actively shapes the visual representation in \model{}, unlike cross-modal encoders where no cross-modal interaction influencing the nature of the vision features occurs.

\begin{table}[t]
\caption{\textbf{Per-scene performance on the CORE benchmark.} \model{} performs consistently best across all indoor and outdoor scenes when using the correct prompt (\checkmark). When text-conditioning on an object other than the one to retrieve (\xmark; incorrect prompt), performance drops significantly. 
}
\label{tab:knn_exp_quant_full}

\centering
\footnotesize
\setlength{\tabcolsep}{3.5pt}
\sisetup{
  table-format=2.1,
  table-number-alignment=center
}
\resizebox{1.0\textwidth}{!}{
\begin{threeparttable}
\begin{tabular}{
l
S S @{\hspace{7pt}}
S S @{\hspace{7pt}}
S S @{\hspace{7pt}}
S S @{\hspace{7pt}}
S S @{\hspace{7pt}}
S S
}
\toprule
\multicolumn{1}{c}{\textbf{Method}} &
\multicolumn{2}{c}{\textbf{Bathroom}} &
\multicolumn{2}{c}{\textbf{Kitchen}} &
\multicolumn{2}{c}{\textbf{Liv. Room}} &
\multicolumn{2}{c}{\textbf{Park}} &
\multicolumn{2}{c}{\textbf{Suburb}} &
\multicolumn{2}{c}{\textbf{Street}} \\
\cmidrule(lr){2-3}\cmidrule(lr){4-5}\cmidrule(lr){6-7}
\cmidrule(lr){8-9}\cmidrule(lr){10-11}\cmidrule(lr){12-13}
&
\multicolumn{1}{c}{\checkmark} & \multicolumn{1}{c}{\xmark} &
\multicolumn{1}{c}{\checkmark} & \multicolumn{1}{c}{\xmark} &
\multicolumn{1}{c}{\checkmark} & \multicolumn{1}{c}{\xmark} &
\multicolumn{1}{c}{\checkmark} & \multicolumn{1}{c}{\xmark} &
\multicolumn{1}{c}{\checkmark} & \multicolumn{1}{c}{\xmark} &
\multicolumn{1}{c}{\checkmark} & \multicolumn{1}{c}{\xmark} \\
\midrule
\rowcolor[rgb]{0.70,0.89,0.99}\textbf{MAE} & 22.0 & n/a & 20.0 & n/a & 19.4 & n/a & 25.6 & n/a & 21.7 & n/a & 22.4 & n/a \\
\rowcolor[rgb]{0.70,0.89,0.99}\textbf{DINOv2} & 54.8 & n/a & 38.0 & n/a & 30.8 & n/a & 56.7 & n/a & 34.3 & n/a & 47.6 & n/a \\
\rowcolor[rgb]{0.70,0.89,0.99}\textbf{SigLIP} & 61.2 & 61.6 & 31.0 & 29.4 & 35.2 & 35.0 & 31.6 & 32.6 & 30.3 & 30.9 & 40.4 & 39.8 \\
\rowcolor[rgb]{0.70,0.89,0.99}\textbf{CLIP} & 64.2 & 64.2 & 37.6 & 37.6 & 47.6 & 47.6 & 44.6 & 44.2 & 26.3 & 26.3 & 44.8 & 44.8 \\
\rowcolor[rgb]{0.70,0.89,0.99}\textbf{FLAIR} & 95.6 & 69.0 & 79.6 & 42.2 & 82.0 & 48.0 & 74.4 & 55.8 & 84.6 & 54.3 & 71.4 & 42.2 \\
\rowcolor[rgb]{0.98,0.70,0.65}\textbf{InternVL3-1B} & 61.8 & 54.4 & 35.6 & 28.4 & 43.0 & 32.6 & 50.2 & 42.3 & 42.9 & 33.1 & 48.6 & 45.6 \\
\rowcolor[rgb]{0.98,0.70,0.65}\textbf{InternVL3-2B} & 89.4 & 70.2 & 70.2 & 42.2 & 60.6 & 36.4 & 75.8 & 57.7 & 81.1 & 50.3 & 78.8 & 57.2 \\
\rowcolor[rgb]{0.98,0.70,0.65}\textbf{Qwen3-VL-2B} & 81.6 & 57.6 & 48.8 & 30.8 & 63.8 & 33.6 & 66.5 & 46.1 & 64.0 & 34.3 & 64.8 & 45.4 \\
\rowcolor[rgb]{0.98,0.70,0.65}\textbf{LFM2.5-VL-1.6B} & 85.0 & 73.8 & 56.4 & 42.0 & 64.6 & 49.0 & 66.5 & 48.8 & 72.0 & 51.4 & 73.4 & 64.0 \\
\rowcolor[rgb]{0.99,0.91,0.58}\textbf{GroundingDINO} & 89.4 & 30.2 & 89.6 & 24.8 & 91.0 & 25.2 & 86.5 & 33.5 & 93.1 & 29.7 & 81.6 & 28.6 \\
\rowcolor[rgb]{0.99,0.91,0.58}\textbf{SAM3} & 97.8 & 35.6 & 95.4 & 27.0 & 91.4 & 26.6 & 94.0 & 33.0 & 95.4 & 21.7 & 85.6 & 26.0 \\
\rowcolor{\methodcolor}\textbf{\model{}} & 99.2 & 44.4 & 98.6 & 43.6 & 96.6 & 49.4 & 93.0 & 58.1 & 97.7 & 56.0 & 90.8 & 34.4 \\

\bottomrule
\end{tabular}
\end{threeparttable}
}
\end{table}

\begin{figure}[ht]
\centering
\includegraphics[width=1.0\textwidth]{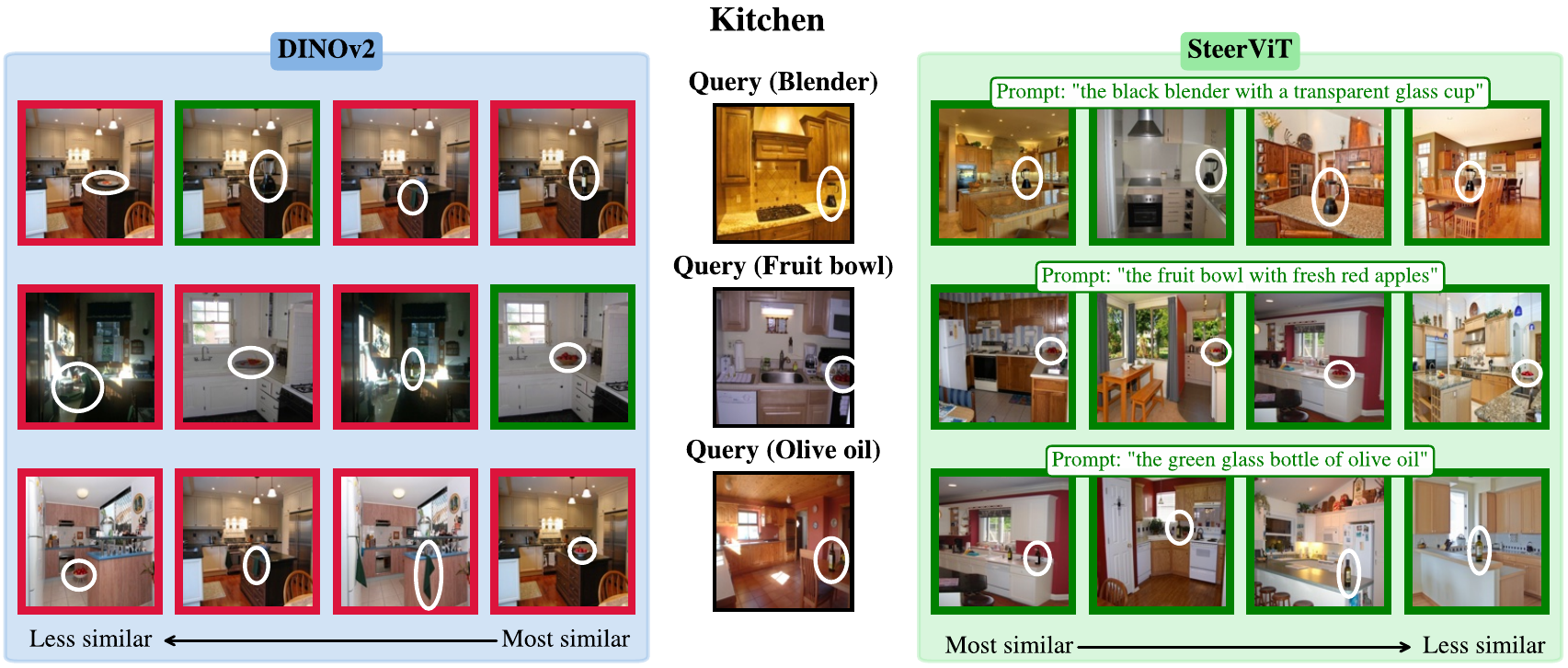}
\includegraphics[width=1.0\textwidth]{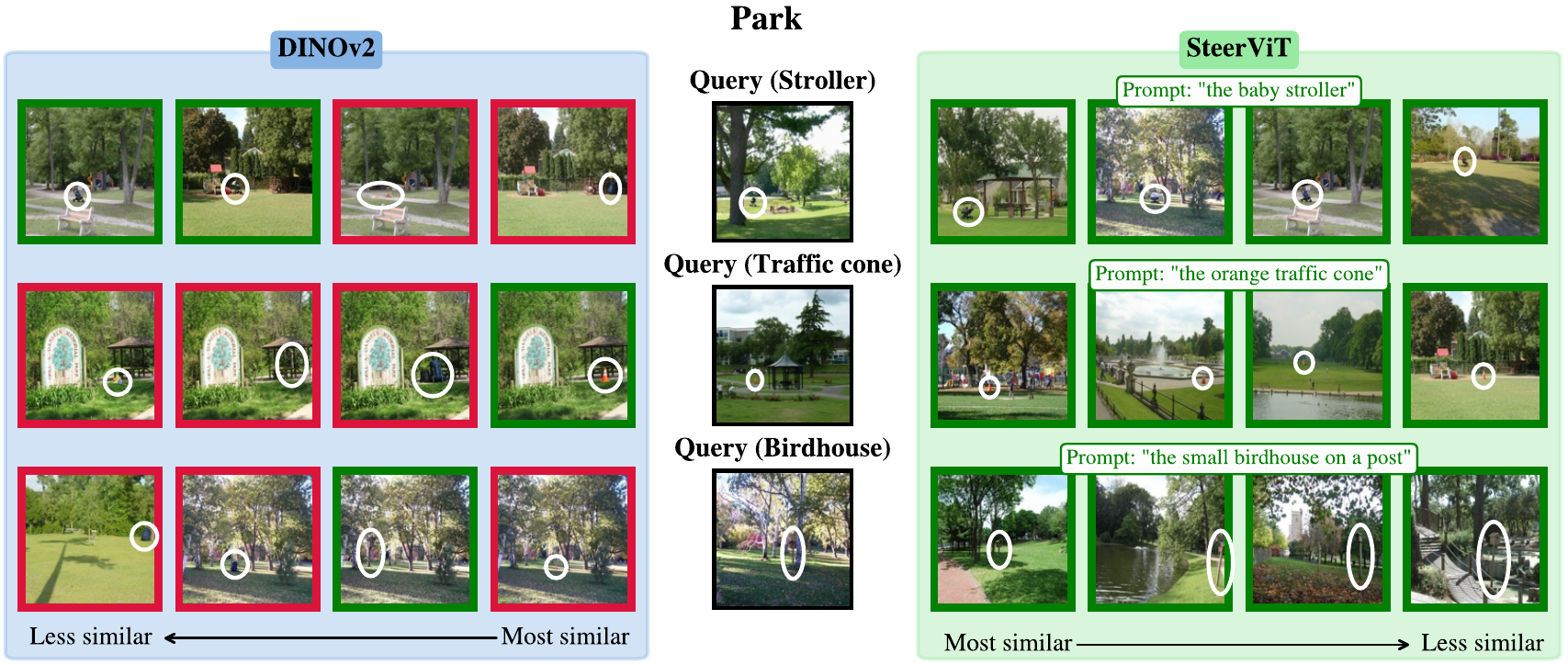}
\vspace{-5mm}
\caption{\textbf{Qualitative image retrieval results on the CORE benchmark.} White circles highlight the inpainted object. \colorbox{OliveGreen}{\color{white}Green}/\colorbox{BrickRed}{\color{white}red} border indicates whether the image contains the queried object of interest. \colorbox{visionblue}{DINOv2} retrieves images mostly based on global scene similarity with little regard for the object of interest. In contrast, \colorbox{\methodcolor}{\model{}} effectively considers the task prompt and ranks images with the correct object highest, despite their inconspicuous background placement and small size.
}
\vspace{-2mm}
\label{apdx_fig:core_qual}
\end{figure}

Figure \ref{apdx_fig:core_qual} illustrates qualitatively how retrievals differ for a prompt-agnostic generic vision model like DINOv2 (blue) and our prompt-aware \model{} (green) for one indoor and outdoor scene each. 
Whereas DINOv2 mostly encodes scene-level appearance and retrieves visually similar images, \model{} focuses on the objects specified in the prompt and retrieves images containing those.
Despite many of the inpainted objects being very small and in the background, the global features incorporate them sufficiently to allow for highly accurate retrievals. 

\subsubsection{Steerability transfers to real-world conditional retrieval.}
\label{apdx:genecis}

\begin{wraptable}{r}{0.38\textwidth}
\centering
\footnotesize
\tabcolsep=1.2mm
\vspace{-10mm}
\caption{%
\textbf{GeneCIS benchmark.}
\model{} outperforms DINOv2 and specialized methods in challenging real-world conditional similarity benchmark.
}
\label{tab:geneCIS}
\begin{tabular}{l ccc}
\toprule
\multirow{2}{*}{Method} & \multicolumn{3}{c}{Focus Object $\uparrow$} \\
& R@1 & R@2 & R@3 \\
\midrule
DINOv2 & 9.6 & 19.5 & 28.4 \\
\rowcolor{\methodcolor}\textbf{\model{}} & \textbf{25.4} & \textbf{39.1} & \textbf{49.9} \\
\color{gray}Specialized & \color{gray}18.7 & \color{gray}30.3 & \color{gray}37.4 \\
\bottomrule
\end{tabular}
\vspace{-6mm}
\end{wraptable}

\begin{figure}[t]
\centering
\includegraphics[width=1.0\textwidth]{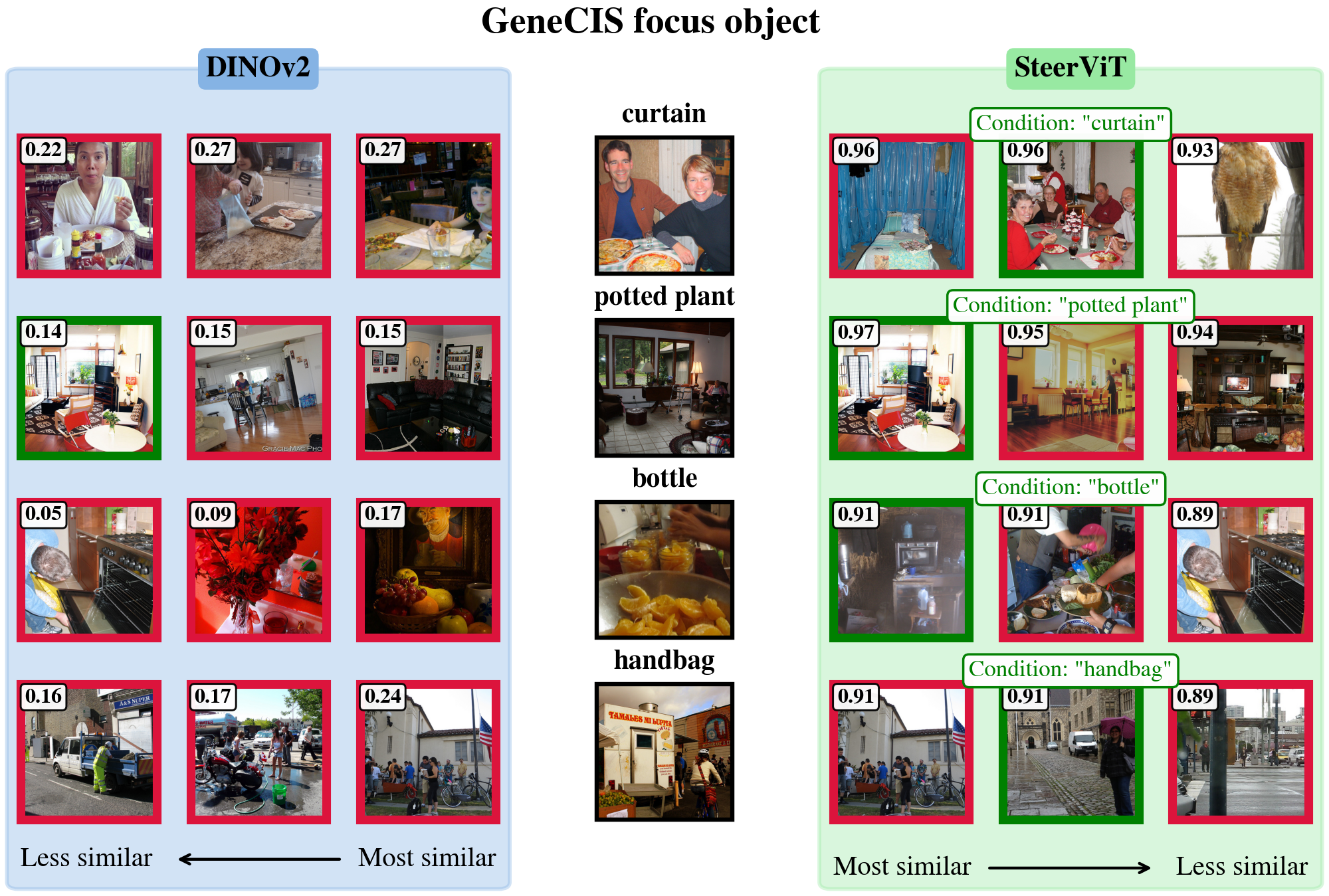}
\caption{
\textbf{Qualitative image retrieval results on the GeneCIS \textit{Focus Object} split.} \colorbox{OliveGreen}{\color{white}Green}/\colorbox{BrickRed}{\color{white}red} border indicates whether images contain the object of interest \textit{and} match the scene context of the query. \colorbox{visionblue}{DINOv2} is biased towards scene similarities whereas \colorbox{\methodcolor}{\model{}} additionally considers the presence of the target object.
}
\label{fig:qual_genecis}
\end{figure}

To assess whether the behavior observed on CORE transfers beyond our controlled inpainting setup, we additionally evaluate on the \textit{Focus Object} split of GeneCIS~\cite{vaze2023gen}, a zero-shot benchmark for conditional image retrieval in real images. Unlike CORE, which explicitly controls object presence and scene context, GeneCIS requires identifying the image that best matches the reference scene while also containing the queried object. The retrieval gallery contains distractors that either share the scene but omit the object or contain the object in a different scene. Despite this more challenging setting, \model{} transfers well in zero-shot evaluation, reaching 25.4\% R@1, compared to 9.6\% for DINOv2 and 18.7\% for the benchmark’s specialized baseline (cf.~\cref{tab:geneCIS}). \cref{fig:qual_genecis} shows the demanding nature of the task and provides qualitative examples comparing DINOv2 rankings with those of \model{}. These results complement CORE and show that language-based representational steering remains effective beyond a controlled synthetic benchmark.

\subsubsection{Layer-wise Effects of Text Conditioning}

To further analyze how text conditioning reshapes the global image representation, we measure the divergence ($1-\cos(Z_v^{(\ell)}, \tilde{Z}_v^{(\ell)})$) between intermediate representations produced by \model{} ($Z_v$) and its underlying vanilla ViT ($\tilde{Z}_v$) after each transformer block $\ell$. Divergence is averaged across all COCO validation images conditioned on a randomly selected object present in the scene.

\begin{figure}[t]
\centering
\includegraphics[width=0.5\textwidth]{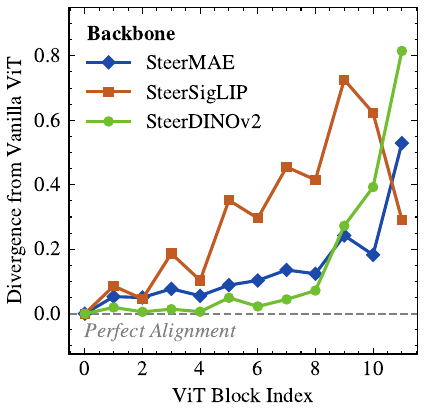}
\caption{
\textbf{Cosine divergence between \model{} and base ViT intermediate features}. Text-conditioning progressively steers internal features across layers.
}
\label{fig:feature_divergence}
\end{figure}

As illustrated in \cref{fig:feature_divergence}, divergence is already non-zero in the early-to-mid layers, indicating that the text signal is incorporated throughout the encoding process. It grows most strongly in later blocks, consistent with the view that earlier ViT layers primarily encode lower-level visual structure while later layers capture higher-level semantics \cite{ghiasi2022visiontransformerslearnvisual}. 
However, the divergence profile is backbone-dependent. SteerDINOv2 remains close to its base model until the final blocks and then departs sharply. 
SteerMAE diverges slowly and gradually throughout
while SteerSigLIP diverges strongly in intermediate layers but converges back toward the original SigLIP space in the final blocks.

\subsection{MOSAIC Benchmark}

The MOSAIC benchmark introduced in \cref{subsec:mosaic_exp} evaluates whether text conditioning can steer the self-attention of the \texttt{[CLS]} token toward patch tokens corresponding to a prompted object.
We stitch four PASCAL-VOC~\cite{pascal_voc} images (padded to $1{:}1$ aspect ratio) into a single $2\times2$ mosaic to prevent a single dominant salient object.
For a given text prompt specifying an object class, the ground-truth is a binary segmentation mask over the mosaic's patch grid, constructed from instances of that class across all four sub-images.

\begin{figure}[ht]
\centering
\includegraphics[width=0.8\textwidth]{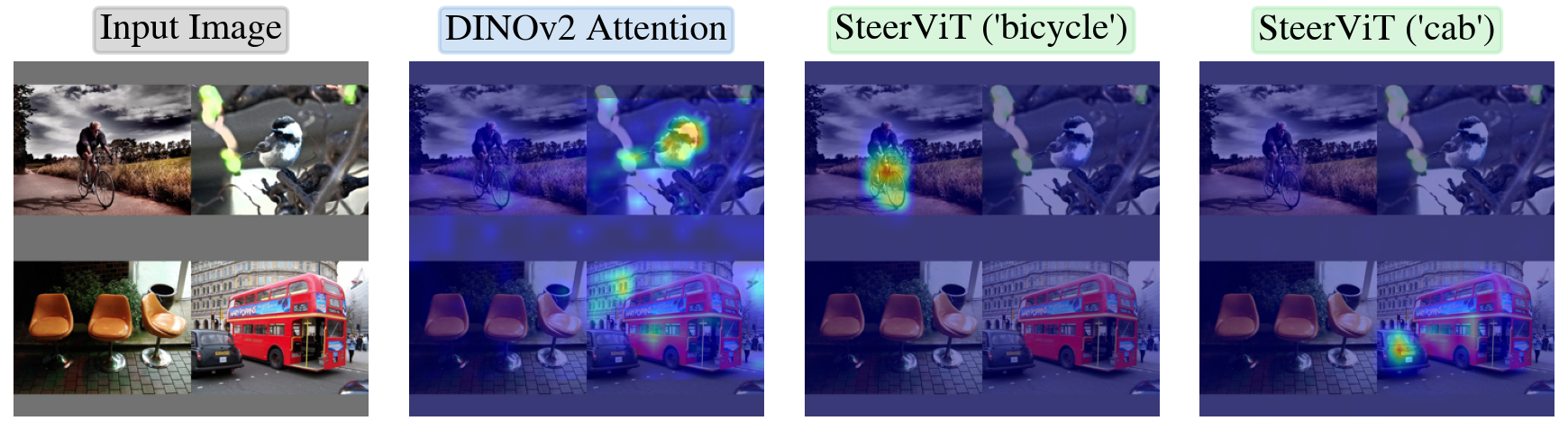}
\includegraphics[width=0.8\textwidth]{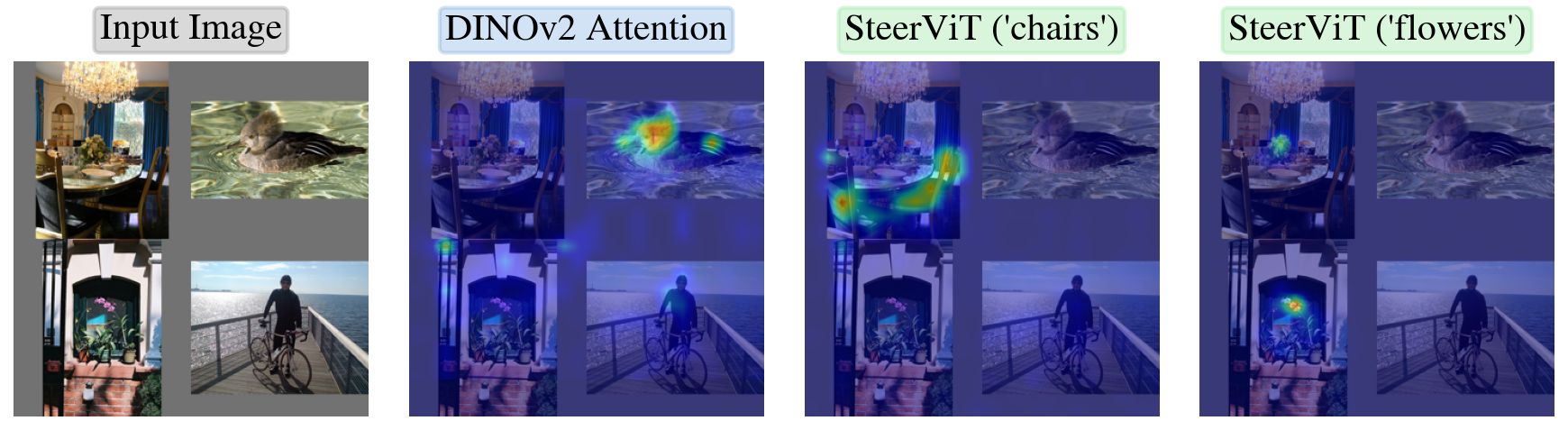}
\includegraphics[width=0.8\textwidth]{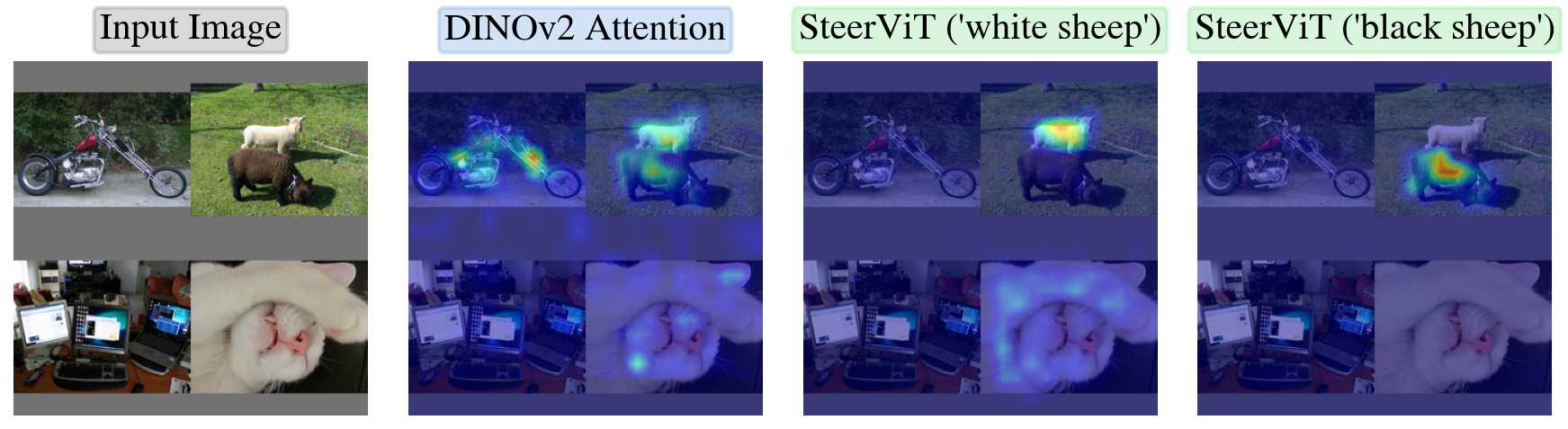}
\vspace{-3mm}
\caption{\textbf{Effective prompt-based attention steering.} \colorbox{visionblue}{DINOv2} focuses on dominant image regions. \textbf{Top:} \colorbox{\methodcolor}{\model{}} can attend to specific objects of interest. \textbf{Middle:} Despite single-instance training, it can consider multiple occurrences of the specified concept. \textbf{Bottom:} specifying attributes (e.g., color) places focus on specific entities. 
}
\vspace{-4mm}
\label{apdx_fig:steer_attn}
\end{figure}

\cref{apdx_fig:steer_attn} provides more qualitative examples of attention maps on such multi-image mosaics. 
Again, we see a strong propensity of DINOv2 to focus on the most dominant object(s) within the collage. \model{} selectively attends to the objects or entities specified in the prompt (\cref{apdx_fig:steer_attn}, top). Despite being trained with single-instance data (i.e., only one object corresponding to the prompt), we find that our model can localize multiple instances of an object class across images in the mosaic (\cref{apdx_fig:steer_attn}, \model{} attends to all instances of ``chair'' in middle row).
We discuss this emergent multi-localization property of our method in ~\cref{app:training_signal}.
\model{} also considers described attributes and routes global attention accordingly (e.g., ``black sheep'' vs. ``white sheep'' in \cref{apdx_fig:steer_attn} (bottom)).

\subsection{Anomaly Segmentation}
\label{apdx:anomaly_segmentation}

\begin{figure}[t]
\centering
\includegraphics[width=1.\textwidth]{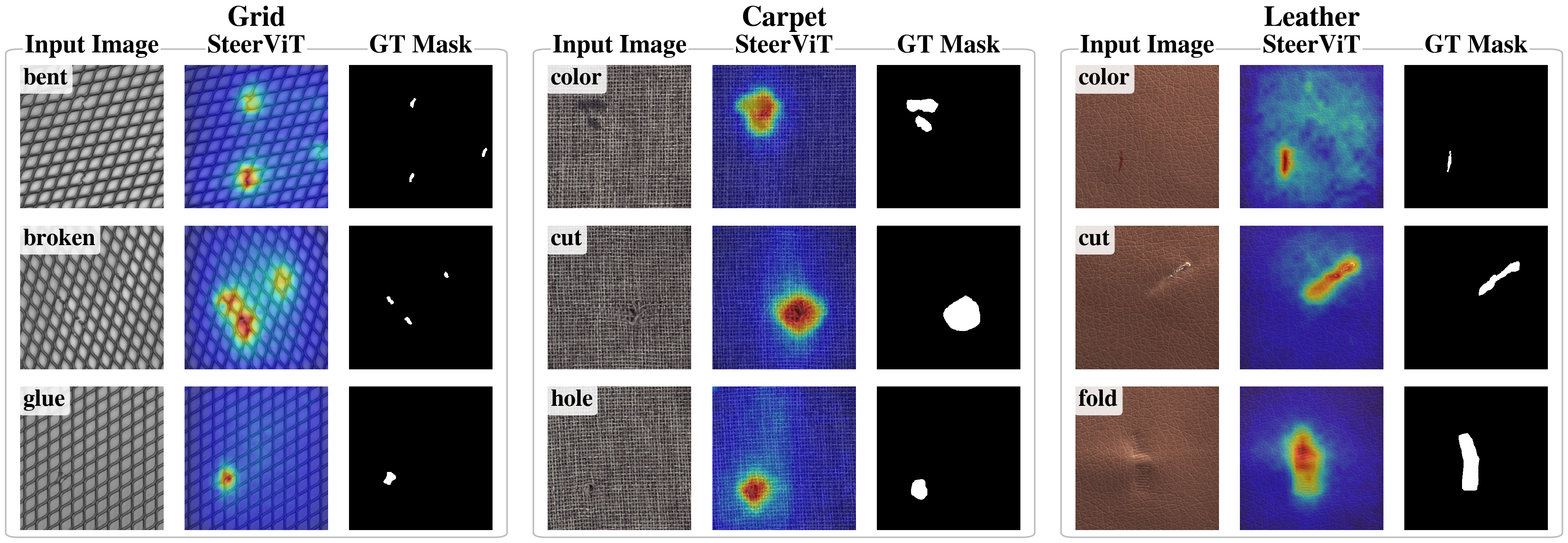}
\includegraphics[width=1.\textwidth]{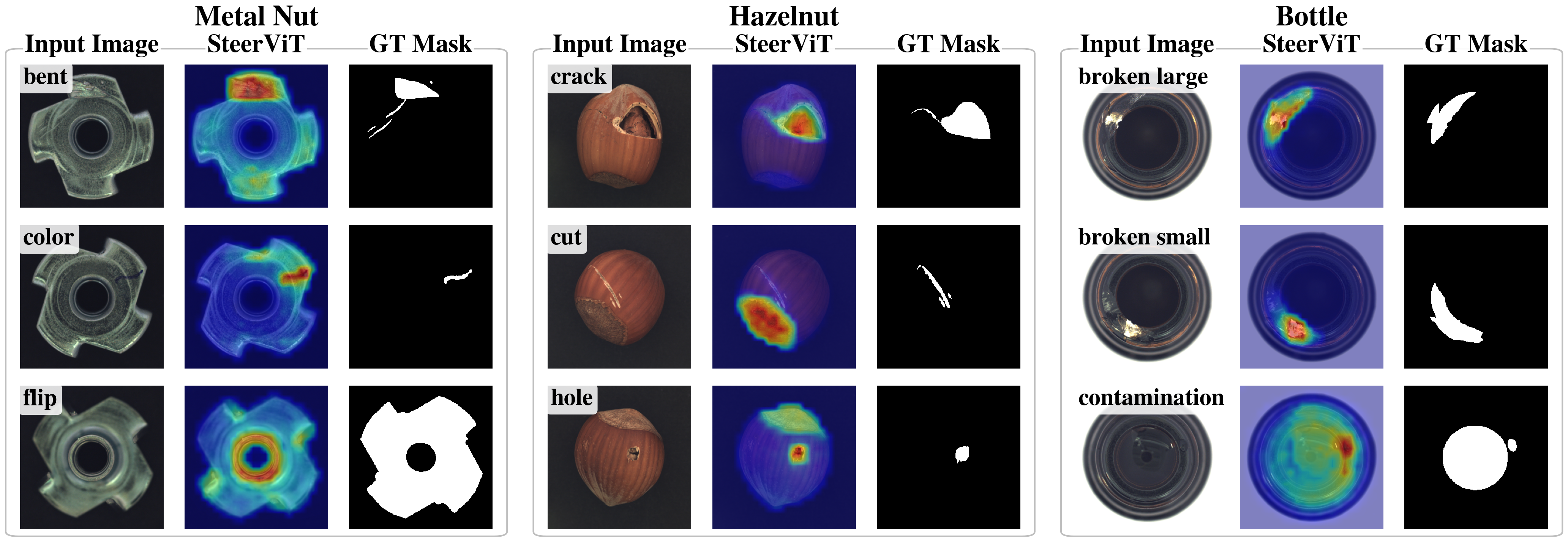}
\vspace{-3mm}
\caption{\textbf{Anomaly segmentation heatmaps predicted by \model{}.} We use the linear classification head optimized for patch-wise referential segmentation. Despite the stark task and domain gap, performance is especially good on texture-based classes (top) but also allows \model{} to highlight small defects in objects (bottom).
}
\vspace{-4mm}
\label{apdx_fig:anomaly_detection_examples}
\end{figure}

We perform zero-shot anomaly segmentation (AS) on the MVTec~AD~\cite{bergmann_mvtech} and VisA~\cite{zou2022_visa_dataset} datasets.
We derive the anomaly map by upsampling the continuous heatmap produced by the learned linear segmentation head.
For all text-steerable methods, we use an ensemble of ten anomaly prompts (e.g., \texttt{``the anomaly in the <object>.''}) and average the resulting heatmaps across prompts.
Following standard practice in the AD literature~\cite{WinCLIP_Jeong23CVPR}, we measure the pixel-wise \textit{area under the ROC curve} ($\text{ROC}_P$), the \textit{Per-Region-Overlap} (PRO), and threshold-optimal $F_1$ score ($F_1^{\max}$).
Performance of existing training- and adaptation-free AS baselines is reported as stated in the original publications.

\begin{table}[th]
\centering
\footnotesize
\setlength{\tabcolsep}{4pt}
\begin{threeparttable}
\vspace{5mm}
\caption{Zero-shot segmentation performance on industrial AD data.}
\label{tab:mvtec_visa}
\vspace{-5mm}
\begin{tabular}{l l c
                S[table-format=2.1] S[table-format=2.1] S[table-format=2.1]
                S[table-format=2.1] S[table-format=2.1] S[table-format=2.1]}
\toprule
\multicolumn{3}{c}{\textbf{Method}} &
\multicolumn{3}{c}{\textbf{MVTec AD}} &
\multicolumn{3}{c}{\textbf{VisA}} \\
\cmidrule(lr){1-3}\cmidrule(lr){4-6}\cmidrule(lr){7-9}
Name & Ref. & AS &
{$\text{ROC}_P$} & {PRO} & {$F_1^{\max}$} &
{$\text{ROC}_P$} & {PRO} & {$F_1^{\max}$} \\
\midrule
MaskCLIP & \cite{zhou2022_maskclip} & \xmark & 63.7    & 40.5    & 18.5   & 60.9   & 27.3   & 7.3    \\
CLIPseg  & \cite{lueddecke22_clipseg} & \xmark & 69.0    & 34.6    & 12.5   & 89.5   & 62.4   & 13.9   \\
SAM3     & \cite{carion2025_sam3} & \xmark & 79.9    & 54.5    & 24.1   & 89.8   & 65.9   & 15.5   \\
\addlinespace
\color{gray}WinCLIP  & \color{gray}\cite{WinCLIP_Jeong23CVPR} & \color{gray}\cmark & \color{gray}85.1    & \color{gray}64.6    & \color{gray}31.7   & \color{gray}79.6   & \color{gray}59.8   & \color{gray}14.8   \\
\color{gray} DIVAD    & \color{gray}\cite{hicsonmez2026training}& \color{gray} \cmark & \color{gray}\underline{88.0}    & \color{gray}73.3    & \color{gray}35.5   & \color{gray}\textbf{93.4}   & \color{gray}78.2   & \color{gray}\textbf{24.0}   \\
\color{gray}FADE     & \color{gray}\cite{li2024_fade} & \color{gray}\cmark & \color{gray}\textbf{89.6}    & \color{gray}\textbf{84.5}    & \color{gray}\textbf{39.8}   & \color{gray}91.5   & \color{gray}\underline{79.3}   & \color{gray}16.7   \\
\addlinespace
\rowcolor{\methodcolor}\textbf{\model{}} & \textbf{Ours} & \xmark & 87.8 & \underline{82.1} & \underline{35.6} & \underline{92.1} & \textbf{82.0} & \underline{18.3} \\
\bottomrule
\end{tabular}
\begin{tablenotes}[flushleft]
\scriptsize
\item AS indicates dedicated anomaly segmentation methods.
\end{tablenotes}
\end{threeparttable}
\end{table}

\model{} again matches dedicated methods on this extreme OOD setting.
As shown in \cref{tab:mvtec_visa}, \model{} reaches 82.1 PRO on MVTec~AD, substantially outperforming off-the-shelf segmentation-based approaches (SAM3 at 54.5, CLIPseg at 34.6) and closing much of the gap to specialist methods such as FADE (84.5).
On VisA, \model{} surpasses FADE in $\text{ROC}_P$ (92.1 vs. 91.5) and PRO (82.0 vs. 79.3), indicating robust transfer to a harder, more diverse inspection setting.
Qualitative examples of \model{}'s predictions on a subset of object types and defect categories are provided in \cref{apdx_fig:anomaly_detection_examples}.
While the predicted heatmaps are especially accurate for texture-based inputs, the zero-shot setting makes it impossible to accurately predict certain defects (e.g., flipped metal nut), as no training on non-anomalous instances occurs and no visible defects (e.g., scratches) are present. 

\subsection{Prompt Complexity}

\begin{figure}[t]
\centering
\includegraphics[width=0.66\linewidth, trim=0 25 0 0, clip]{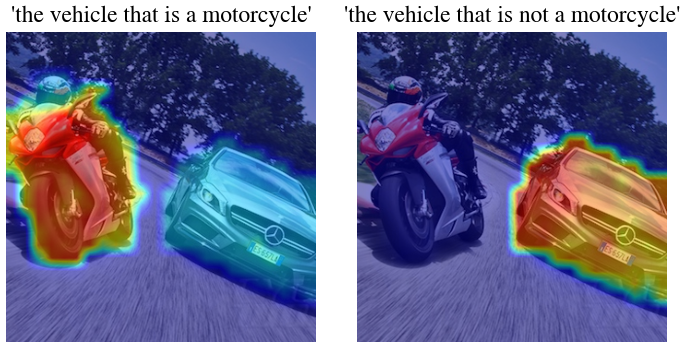}
\hfill
\includegraphics[width=0.315\linewidth]{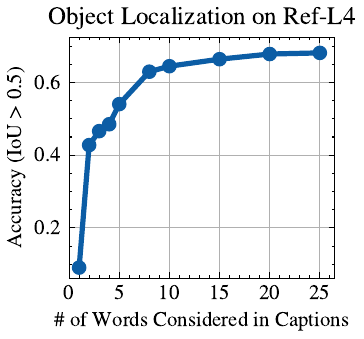}
\vspace{-0.5mm}
\caption{Prompt Complexity: Negation changes the localized object (left). More detailed descriptions improve localization performance (right). \model{} generalizes beyond simple object-referring prompts.}
\label{fig:negation_token_length}
\end{figure}

With the exception of the detailed object descriptions in PODS, the majority of prompts during text-conditioned visual feature extraction are short in length and simple in nature. \cref{fig:negation_token_length} shows how \model{} deals with more complex constructs beyond short object names. On the left, we show that nuanced linguistic constructs such as negation can also be handled by the model. On the right, we pre-filter the detailed referring expressions of the Ref-L4~\cite{peng2024refl4} dataset to instances with at least 30 words, gradually vary the length of the prompts provided to \model{} and measure the referring object localization accuracy. The resulting graph shows that performance improves as compositionally richer and more information-dense text-conditioning is applied, indicating that the model is able to beneficially incorporate additional details and can generalize beyond short and simple prompts.

\section{Additional Analysis on Training Design Choices}
\label{apdx:more_ablations}

Below, we further analyze important design choices involved in the training of \model{}. To gauge model performance, we report the preservation of representation quality, as measured by fine-grained linear classification probes (FG-CLS) and binary object of interest segmentation (ADE20k), using the previously introduced methodology. Models' steerability and textual understanding are assessed using the CORE and PODS benchmarks. We also evaluate referential grounding on RefCOCOg by computing IoU between the predicted patch-level segmentation mask and the ground-truth mask.

\subsection{Training Objective: Pointing vs.\ Segmenting}
\label{app:training_signal}

Our goal is to instill language understanding into a frozen ViT via a simple, scalable proxy task.
Referential localization serves this purpose: given a text prompt, the model must 
ground the described object.
To avoid the complexity of a pixel-level decoder, we operate entirely in the ViT's $n \times n$ patch-token grid.

\paragraph{Pointing.}
The simplest formulation assigns a hard target (1 at the center patch, 0 elsewhere), but we find this unstable to optimize in practice for both classification and regression.
Instead, we pass a Gaussian kernel ($\sigma{=}1.1$) at the bounding-box center and train with soft cross-entropy, which provides a smooth gradient towards the object center.
However, this target is invariant to object shape and size, collapsing supervision to a single spatial location.

\paragraph{Segmenting.}
Consequently, we generate segmentation masks using SAM2~\cite{ravi2024sam2} conditioned on ground-truth bounding boxes and project them onto the patch grid.
Unlike pointing, this target activates all tokens overlapping the object, teaching content-matching: each patch must determine whether it depicts the described object, not merely how close it is to the center.

\begin{table}[t]
\centering
\footnotesize
\setlength{\tabcolsep}{6pt}
\begin{threeparttable}
\caption{%
\textbf{Role of supervision signal.} 
Segmentation is a superior training objective compared to pointing (Gaussian kernel at bounding box center) across all downstream evaluations, resulting in improved visual and multimodal understanding.
}
\label{tab:supervision}
\begin{tabular}{@{} l
                S[table-format=2.1]
                S[table-format=2.1]
                S[table-format=2.1]
                S[table-format=2.1]@{}}
\toprule
{Train Objective}
  & {FG-CLS $\uparrow$}
  & {ADE20k $\uparrow$}
  & {CORE\ $\uparrow$}
  & {PODS $\uparrow$} \\
\midrule
Pointing & {80.4} & {47.4} & {95.2} & {45.7} \\
\rowcolor{\methodcolor}
Segmentation & \textbf{87.7}  & \textbf{{55.4}} & \textbf{96.0} & \textbf{58.1} \\
\rowcolor{\methodcolor}
\bottomrule
\end{tabular}
\end{threeparttable}
\end{table}

As shown in \cref{tab:supervision}, the segmentation objective consistently outperforms pointing across all metrics.
The largest gains appear on feature quality preservation (FG-CLS: +7.3; ADE20k: +8.0) and PODS (+12.4). Pointing teaches the model \textit{where} an object is, whereas segmentation also teaches \textit{what} the object looks like, producing features that better encode object extent and appearance.

\subsection{Training Duration}
\label{subsec:training_duration}

Steerability emerges rapidly: within 50k iterations of training, CORE accuracy reaches 95.3\%
(vs.\ 43.7 for frozen DINOv2), 
while FG-CLS remains nearly constant at 89.6 (vs.\ 89.0 for frozen DINOv2).
However, tasks that require deeper language understanding continue improving with longer training: between 50k and 450k iterations, 
PODS improves from 49.9 to 58.1
and RefCOCOg from 63.4 to 70.6, suggesting that longer training instills richer multimodal representations rather than merely teaching the model to route attention.
We adopt 500k iterations for all final models.

\subsection{Scaling}
\label{subsec:scaling}
\begin{table}[t]
\centering
\footnotesize
\setlength{\tabcolsep}{6pt}
\begin{threeparttable}
\caption{%
\textbf{Scaling visual and text encoders.}
Larger ViT backbones improve performance for most tasks.
Reducing RoBERTa-Large to Base modestly degrades visual quality (FG-CLS, ADE20k) but preserves multimodal understanding (CORE, PODS).
}
\label{tab:scaling}
\begin{tabular}{@{} l l 
                S[table-format=2.1]
                S[table-format=2.1]
                S[table-format=2.1]
                S[table-format=2.1]@{}}
\toprule
\rowcolor{white}{ViT} & {RoBERTa} 
  & {FG-CLS $\uparrow$}
  & {ADE20k $\uparrow$}
  & {CORE $\uparrow$}
  & {PODS $\uparrow$} \\
\midrule
Small & Large & {80.0} & {50.8} & {93.6} & {44.1} \\
\rowcolor{\methodcolor}
Base & Large & \textbf{87.7} & {55.4} & {96.0} & {58.1} \\
Large & Large & {85.8}  & \textbf{55.5} & \textbf{96.8} & \textbf{62.8} \\
\addlinespace
\midrule
\addlinespace
Base & Base & {\textbf{87.9}}  & {53.6} & {95.7} &  {57.4} \\
\rowcolor{\methodcolor}
Base & Large & {87.7} & \textbf{55.4} & \textbf{96.0} & \textbf{58.1} \\
\rowcolor{\methodcolor}
\bottomrule
\end{tabular}
\end{threeparttable}
\end{table}

As shown in \cref{tab:scaling}, scaling the vision backbone from ViT-S through ViT-B to ViT-L leads to improvements in both representational quality and textual understanding. 
Interestingly, scaling the text encoder from RoBERTa-Base (125M parameters) to  RoBERTa-Large (355M) also improves performance on visual tasks. 
This suggests that rich embeddings of the larger text encoder add more semantic structure to the ViT's residual stream (row 4 vs row 5).

\model{} exhibits considerable robustness to the amount of training data utilized. As shown in \cref{tab:data_ablation}, even when using just a fourth of the unique samples during training, steerability and feature fidelity remain high.

\begin{table}[t]
\centering
\footnotesize
\caption{
Ablating dataset size with models trained for 200k iterations.
}
\label{tab:data_ablation}
\begin{tabular}{l c c c c c}
\toprule
\cellcolor{white}\textbf{Setting}
& FG-CLS
& ADE20k
& CORE
& PODS
& Avg. \\
\midrule
Data 25\%     & 88.4 & 53.2 & 95.3 & 55.6 & 73.1 \\
Data 50\%      & 88.0 & 54.0 & 95.2 & 52.8 & 72.5 \\
\rowcolor{\methodcolor}
Default        & \textbf{89.5} & \textbf{55.3} & \textbf{95.5} & 55.4 & 73.9 \\
\bottomrule
\end{tabular}%
\end{table}

\subsection{Role of FFN}

\begin{table}[ht]
\centering
\footnotesize
\setlength{\tabcolsep}{6pt}
\begin{threeparttable}
\caption{%
\textbf{Role of the FFN.}
Removing the gated FFN from each cross-attention block yields comparable or better performance.
The effect is prominent for MAE, where adding FFN degrades steerability and zero-shot transfer substantially.
}
\label{tab:ffn_ablation}
\begin{tabular}{l c
                S[table-format=2.1]
                S[table-format=2.1]
                S[table-format=2.1]
                S[table-format=2.1]
                S[table-format=2.1]}
\toprule
\rowcolor{white}{Backbone} & {FFN}
  & {FG-CLS $\uparrow$}
  & {ADE20k $\uparrow$}
  & {CORE\ $\uparrow$}
  & {PODS\ $\uparrow$}
  & {MVTec $\uparrow$} \\
\midrule
\rowcolor{\methodcolor}DINOv2 & \xmark & \textbf{87.7} & \textbf{55.4} & \textbf{96.0} & \textbf{58.1} & \textbf{82.1} \\
DINOv2 & \cmark & {86.0} & {54.3} & 95.0 & {55.2} & 79.3 \\
\addlinespace
\rowcolor{\methodcolor}SigLIP & \xmark & \textbf{82.6} & {47.7} & \textbf{91.3} & \textbf{27.4} & 74.8 \\
SigLIP & \cmark & 80.5 & \textbf{48.5} & 89.1 & 26.9 & \textbf{80.8} \\
\addlinespace
\rowcolor{\methodcolor}MAE & \xmark & {67.3} & \textbf{35.9} & \textbf{74.9} & \textbf{23.8} & \textbf{77.3} \\
MAE & \cmark & \textbf{67.8} & 35.5 & 67.7 & 21.8 & 73.9 \\
\bottomrule
\end{tabular}
\end{threeparttable}
\end{table}

\model{} inverts the gated cross-attention formulation introduced by the Flamingo MLLM~\cite{alayrac2022flamingo}. However, it forgoes secondary gated feed-forward networks (FFN) following the cross-attention operations in the original architecture. As shown in \cref{tab:ffn_ablation}, including the FFN brings little to no benefit in representation quality, while consistently hurting steerability and out-of-distribution transfer. The effect is especially pronounced for MAE, where adding the FFN reduces CORE by 7.2 points and MVTec PRO by 3.4 points (row 5 vs.\ row 6). At the same time, the FFN substantially increases the parameter count of the adapter, growing the cross-attention module from 21.2M additional parameters without a FFN to 35.4M with FFN (\mbox{+67\%}). Since the FFN is both expensive and empirically dispensable, we omit it in the final architecture.

\section{Further Details on Experimental Setup}
\label{apdx:exp_details}

\subsection{Training Data}
\label{apdx:data}

\begin{wrapfigure}{r}{0.4\textwidth}
\centering
\includegraphics[width=\linewidth]{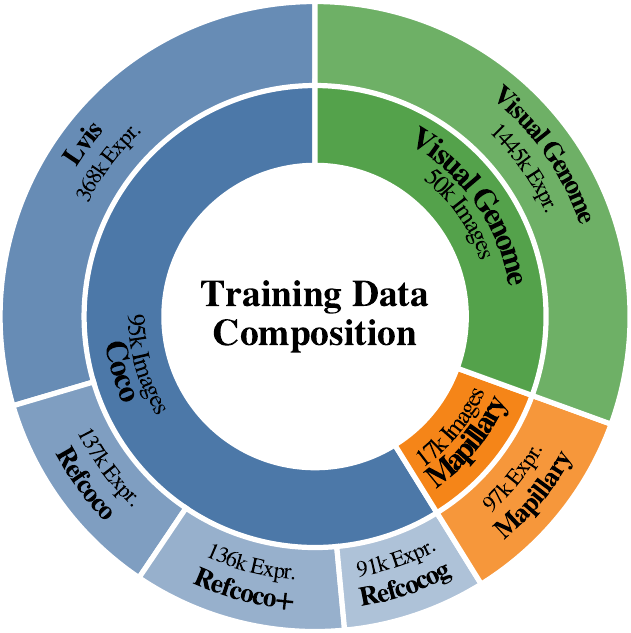}
\caption{\textbf{Training data.} The inner ring shows unique images per source while the outer ring shows associated referring expressions. %
}
\vspace{-5mm}
\label{fig:training_data}
\end{wrapfigure}

We train on a mixture of referential segmentation and grounding datasets to ensure diversity in both visual domains and textual expression styles (see Fig.~\ref{fig:training_data}):

\noindent \textbf{RefCOCO, RefCOCO+, RefCOCOg}~\cite{kazemzadeh2014referitgame, yu2016refcocog} provide referring expressions grounded in COCO images.
RefCOCO+ excludes spatial language (e.g., ``left of''), encouraging the model to rely on appearance cues, while RefCOCOg contains longer, more descriptive expressions that exercise the model's capacity for detailed textual understanding.

\noindent \textbf{LVIS}~\cite{gupta2019lvis} uses the same underlying COCO images but also considers fine-grained and long-tail object categories.

\noindent \textbf{Visual Genome}~\cite{krishna2017visual} (preprocessed following MDETR~\cite{kamath2021mdetr}) contributes region descriptions paired with bounding boxes across densely annotated scenes.
These descriptions span a broader vocabulary and more complex spatial relationships than the RefCOCO family, increasing the diversity of text conditioning signals.
We use SAM2~\cite{ravi2024sam2} to convert bounding boxes to binary segmentation masks.

\noindent \textbf{Mapillary Vistas}~\cite{neuhold2017mapillary} provides street-level imagery with fine-grained panoptic annotations.
Including this dataset expands the visual domain beyond COCO,
improving out-of-distribution generalization.
We adopt Describe Anything's~\cite{lian2025describeanything} synthetic
referential expressions along with accompanying segmentation masks.

Overall, our combined dataset of 162k unique images and 2.28M image-text pairs exposes the model to varied scene complexities (from single objects to dense urban panoramas), expression lengths (from two-word labels to multi-sentence descriptions), and visual domains (indoor, outdoor, street-level), encouraging robust steered representations.

\subsection{Baseline Feature Extraction}
\label{app:baseline_extraction}

This section provides more details on how we extract text-conditioned (or text-fused) visual features from each baseline family evaluated in \cref{sec:experiments}.
For all baselines, global feature aggregation follows the convention of the base method (e.g., \texttt{[CLS]} token or mean pooling) unless stated otherwise.

\paragraph{Cross-Modal Encoders (CLIP \cite{radford2021_clip}, SigLIP \cite{zhai2023_siglip}).}
Images and language inputs are embedded independently by their respective encoders.
We combine the resulting visual and text feature vectors via element-wise addition (late fusion), the simplest form of post-hoc multimodal interaction.

\paragraph{MLLMs (InternVL3~\cite{zhu2025internvl3}, Qwen3-VL~\cite{Qwen3-VL}).}
We feed the text prompt~$X_t$ and image~$X_v$ as a single multimodal input sequence and extract hidden states from the last LLM layer.
For dense features, we take the hidden state at each image-token position.
For global features, following E5-V~\cite{Jiang2024E5VUE}, we append an instruction asking the model to summarize the image in one word in addition to the initial text prompt $X_t$
and extract the last-token hidden state.

\paragraph{Open-Vocabulary Localization (SAM3~\cite{carion2025_sam3}, GroundingDINO\cite{liu2023groundingdino}).}
We ignore decoder outputs and extract intermediate features from the multimodal encoder.
For SAM3, we take the patch-level outputs of the \textit{Fusion Encoder} and apply mean pooling to obtain global image features.
For GroundingDINO, we use the outputs of the cross-modal \textit{Feature Enhancer} layer.
Because features are available at multiple spatial scales, we first interpolate them to a common intermediate resolution before averaging to obtain dense pseudo-patch-level features that are optionally mean-pooled to yield global image-level embeddings.

\subsection{Assessing Visual Representation Quality}
\label{apdx:feature_quality}

In addition to the steerability of vision representations, we also evaluate their quality for common computer vision downstream tasks. 

\noindent\textbf{Fine-grained image classification} tests whether representations capture sufficient detail to perform accurate categorization within specialized domains. Namely, we train a linear probe (300 epochs, LR: $1e^{-3}$, BS: 128) on frozen features extracted from the ImageWoof~\cite{Howard_Imagewoof_2019}, Waterbirds~\cite{Sagawa2020_waterbirds}, and StanfordCars~\cite{stanford_cars} datasets. 
Here, prompt-aware models are conditioned on \texttt{``the dog''}, \texttt{``the bird''}, and \texttt{``the car''}, respectively. 
\newline
\noindent\textbf{Binary object-of-interest segmentation} tests the semantic quality of patch-level features using the ADE20k~\cite{zhou2017scene} dataset.
We apply the training-free OpenHummingBird~\cite{pariza2024hbird} evaluation methodology where patch-class pairs are precomputed for the training split and patch-wise segmentation is performed at inference via nearest-neighbor retrieval from the reference bank. Each class is evaluated separately (i.e., binary class/non-class segmentation objective) and features are conditioned on the object of interest's original class name (e.g., \texttt{``the chair''}). 
\newline
\noindent\textbf{Action recognition} tests whether image representations aggregated over time capture the action performed in a video. Namely, for the UCF101~\cite{soomro2012arxiv} and HMDB51~\cite{kuehne2011hmdb} datasets, we uniformly sample 8 frames per video, extract frozen frame-level features conditioned on \texttt{``the action performed by the person''}, $\ell_2$-normalize them before and after temporal averaging, and train a logistic-regression linear probe on the resulting video-level representations.
\newline\newline
We provide additional results on fine-grained classification, binary object-of-interest segmentation as well as action recognition in \cref{tab:feature_preservation}. 
\model{} maintains 98.8\% of DINOv2's average performance across these tasks, further emphasizing the high degree of feature quality preservation for various downstream tasks as indicated in the main paper.

\begin{table}[t]
\centering
\footnotesize
\setlength{\tabcolsep}{3.5pt}
\renewcommand{\arraystretch}{0.92}

\caption{
SteerViT maintains 98.8\% of DINOv2's avg. performance across
\textcolor{clstext}{classification},
\textcolor{segtext}{segmentation}, and
\textcolor{acttext}{action recognition} downstream tasks.
}
\vspace{-3mm}
\label{tab:feature_preservation}
\resizebox{\linewidth}{!}{
\begin{tabular}{@{}l
>{\columncolor{clsbg}}c
>{\columncolor{clsbg}}c
>{\columncolor{clsbg}}c
>{\columncolor{segbg}}c
>{\columncolor{segbg}}c
>{\columncolor{actbg}}c
>{\columncolor{actbg}}c c@{}}
\toprule
\textbf{Model}
& \textcolor{clstext}{Birds}
& \textcolor{clstext}{Cars}
& \textcolor{clstext}{Dogs}
& \textcolor{segtext}{ADE20k}
& \textcolor{segtext}{City.}
& \textcolor{acttext}{UCF101}
& \textcolor{acttext}{HMDB51} & Avg. \\
\midrule
DINOv2                  & \textbf{94.7} & \textbf{83.8} & 89.1 & 53.7 & 52.8 & \textbf{92.3} & \textbf{62.3}  & \textbf{88.1}\\
SteerViT & 91.2 & 77.7 & \textbf{94.3} & \textbf{55.4} & \textbf{53.8} & 89.4 & 60.0  & 87.0 \\
\bottomrule
\vspace{-4mm}
\end{tabular}}
\end{table}

\begin{table}[t]
\centering
\caption{
Efficiency during \textcolor{traintext}{multimodal training} and
\textcolor{infertext}{inference}.
Average over 256 images (336x336; batch size 16) with 20-word prompts (A100 GPU).  %
}
\vspace{-3mm}
\label{tab:efficiency}
\begin{tabular}{l c
>{\columncolor{trainbg}}c
>{\columncolor{trainbg}}c
>{\columncolor{inferbg}}r
>{\columncolor{inferbg}}r
>{\columncolor{inferbg}}r}
\toprule
Name & Params
& \textcolor{traintext}{Tuned}
& \textcolor{traintext}{Samples}
& \shortstack[c]{\textcolor{infertext}{Compute}\\[-1pt]{\scriptsize GFLOPS/img}}
& \shortstack[c]{\textcolor{infertext}{Latency}\\[-1pt]{\scriptsize ms/img}}
& \shortstack[c]{\textcolor{infertext}{Memory}\\[-1pt]{\scriptsize peak GiB}} \\
\midrule
DINOv2        & 86M  & n/a & n/a   & 98.6   & 6.6  & 1.0  \\
\rowcolor{\methodcolor}
SteerViT      & 465M & Gated CA & 2.28M & 133.8  & 9.5  & 3.0  \\
SAM3          & 840M & all & 146M  & 556.0  & 35.1 & 5.3  \\
InternVL3-2B  & 2.1B & all & n/r   & 1591.0 & 12.2 & 6.3  \\
\bottomrule
\vspace{-6mm}
\end{tabular}%
\end{table}

\subsection{Analyzing and Comparing Compute Efficiency}

The resource requirements of \model{} and selected baselines are reported in \cref{tab:efficiency}. 
As \model{} only trains 21.2M CA parameters and keeps the underlying ViT and text encoder frozen, training is substantially lighter than baselines. %

\end{document}